\documentclass[10pt,twocolumn,letterpaper]{article}

\usepackage{iccv}
\usepackage{times}
\usepackage{epsfig}
\usepackage{graphicx}
\usepackage{amsmath}
\usepackage{amssymb}

\usepackage{mathtools}
\usepackage{multicol}

\usepackage{threeparttablex}
\usepackage{longtable}
\usepackage[dvipsnames]{xcolor}

\usepackage{scalerel}
\usepackage{stmaryrd}
\usepackage{multirow}
\usepackage{subcaption}
\usepackage{float}
\usepackage{wrapfig}
\usepackage{cite}
\usepackage{makecell}
\usepackage{enumitem}
\usepackage{booktabs} 
\usepackage[font=small]{caption} 
\usepackage{sidecap}
\sidecaptionvpos{figure}{c}

\usepackage{subcaption}
\usepackage{graphicx}
\usepackage{lipsum}
\usepackage[normalem]{ulem}

\usepackage{pifont}
\usepackage{listings}
\usepackage{xcolor}
\usepackage[export]{adjustbox}

\definecolor{codegreen}{rgb}{0,0.6,0}
\definecolor{codegray}{rgb}{0.5,0.5,0.5}
\definecolor{codepurple}{rgb}{0.58,0,0.82}
\definecolor{backcolour}{rgb}{0.95,0.95,0.92}

\lstdefinestyle{mystyle}{
	backgroundcolor=\color{backcolour},
	commentstyle=\color{codegreen},
	keywordstyle=\color{magenta},
	numberstyle=\tiny\color{codegray},
	stringstyle=\color{codepurple},
	basicstyle=\ttfamily\footnotesize,
	breakatwhitespace=false,
	breaklines=true,
	captionpos=b,
	keepspaces=false,
	numbers=left,
	numbersep=5pt,
	showspaces=false,
	showstringspaces=false,
	showtabs=false,
	tabsize=2,
	basicstyle=\ttfamily\scriptsize
}

\usepackage[pagebackref=true,breaklinks=true,letterpaper=true,colorlinks,bookmarks=false]{hyperref}

\definecolor{mycolor}{HTML}{F7F8E0}
\definecolor{darkcyan}{RGB}{0,113,194}
\definecolor{forestgreen}{RGB}{34,139,34}

\hypersetup{colorlinks,breaklinks=true,
	citecolor=darkcyan
}

\newcolumntype{H}{>{\setbox0=\hbox\bgroup}c<{\egroup}@{}}
\newcolumntype{Z}{>{\setbox0=\hbox\bgroup}c<{\egroup}@{\hspace*{-\tabcolsep}}}

\iccvfinalcopy 

\def\iccvPaperID{7308} 

\ificcvfinal\fi

\begin{document}

\title{DocFormerv2: Local Features for Document Understanding}

\author{Srikar Appalaraju$^{1}$ \thanks{Corresponding author.}, \quad Peng Tang$^{1}$ , \quad Qi Dong$^{1}$, \quad Nishant Sankaran$^{1}$, Yichu Zhou$^{2}$ \thanks{Work conducted during an internship at Amazon.}, \quad R. Manmatha$^{1}$ \\
$^{1}$ AWS AI Labs \quad
$^{2}$ School of Computing at University of Utah\\
{\tt\small \{srikara, tangpen, qdon, nishsank, manmatha\}@amazon.com, flyaway@cs.utah.edu}
}

\maketitle
\ificcvfinal\thispagestyle{empty}\fi

\newcommand{\xrightarrowdbl}[2][]{%
	\leftarrow\mathrel{\mkern-14mu}\xrightarrow[#1]{#2}
}
\newcommand{\papertitle}{DocFormerv2 }
\newcommand{\papertitlenospace}{DocFormerv2}
\newcommand{\papertitleshort}{DFv2 }
\newcommand{\papertitleshortnospace}{DFv2}
\renewcommand\theadset{\def\arraystretch{.85}}

\begin{abstract}
   We propose \papertitlenospace, a multi-modal transformer for Visual Document Understanding (VDU). The VDU domain entails understanding documents (beyond mere OCR predictions) e.g., extracting information from a form, VQA for documents and other tasks. VDU is challenging as it needs a model to make sense of multiple modalities (visual, language and spatial) to make a prediction. Our approach, termed \papertitle is an encoder-decoder transformer which takes as input - vision, language and spatial features. \papertitle is pre-trained with unsupervised tasks employed asymmetrically i.e., two novel document tasks on encoder and one on the auto-regressive decoder. The unsupervised tasks have been carefully designed to ensure that the pre-training encourages local-feature alignment between multiple modalities. \papertitle when evaluated on nine  datasets shows state-of-the-art performance over strong baselines e.g. TabFact (4.3\%), InfoVQA (1.4\%), FUNSD (1\%). 
   Furthermore, to show generalization capabilities, on three VQA tasks involving scene-text, \papertitle outperforms previous comparably-sized models and even does better than much larger models (such as GIT2, PaLi and Flamingo) on some tasks.
    Extensive ablations show that due to its pre-training, \papertitle understands multiple modalities better than prior-art in VDU.
\end{abstract}
\vspace{-2em}


\section{Introduction}
\label{sec:intro}

\begin{figure}
  \centering
  \includegraphics[width=1.0\linewidth]{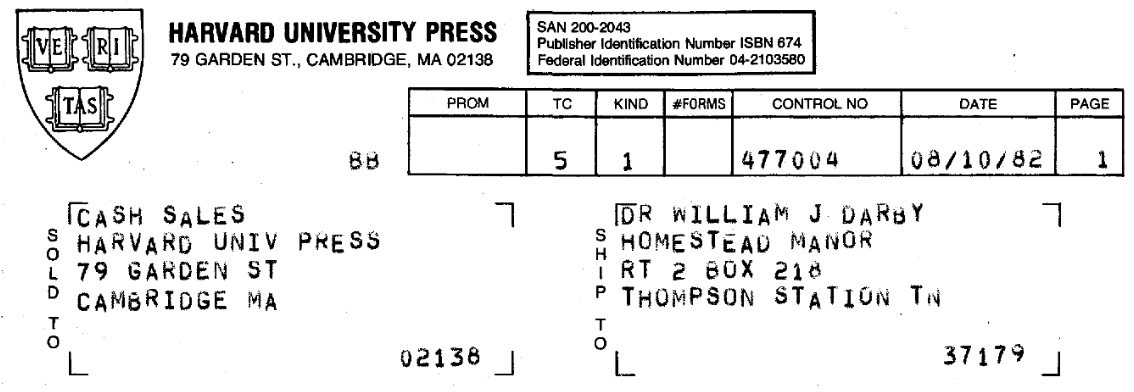}
  \caption{\textbf{Visual Document Understanding}. Snippet of a document receipt from DocVQA \cite{mathew2021docvqa}. VDU tasks could include a model asked to predict "SOLD TO" address (VQA) or predict all relations ("SOLD TO" $\rightarrow$ $<$address$>$, "SHIP TO" $\rightarrow$ $<$address$>$) or asked to infer info from table (at the top).
  }
  \label{fig:splash}
  \vspace{-5mm}
\end{figure}

Documents have become ubiquitous carriers of information, including forms, tables, invoices, and other structured documents. Many such documents require visual and layout understanding to make sense (just the text string is insufficient). Visual Document Understanding (VDU) is the task of leveraging machine learning techniques to comprehend such scanned documents, such as PDFs or images. Popular VDU tasks include Document and Tables VQA \cite{mathew2020docvqa,Chen2019TabFactAL}, sequence labeling for key-value identification in forms \cite{Jaume2019FUNSDAD}, entity extraction \cite{park2019cord}, and document classification \cite{harley2015icdar}. While modern deep-learning based OCR models \cite{litman2020scatter} have proven to be effective in extracting text from documents, the naive approach of linearizing the OCR-text and feeding it to a language model is sub-optimal. This is because the content of a document is presented according to a visual layout and structure that must be taken into account for accurate understanding. Naively linearizing the text from left-to-right will result in sub-optimal performance as the semantic meaning alters based on layout, as shown in Figure \ref{fig:splash} - Table \ref{table:cord},\ref{table:funsd} has experiments demonstrating this. Instead, VDU requires a multi-modal approach that can comprehend text and visual features in the context of a document's 2D layout.

Multi-modal training in general entails feature alignment. Specific to vision-language learning this means aligning a piece of text with an arbitrary span of pixels in visual space \cite{hoyoro,Kim2021ViLTVT,CLIPRadford2021LearningTV,wang2022git,Alayrac2022FlamingoAV,biten2022latr,appalaraju2021docformer,Hao2022MixGenAN,appalaraju2020towards,Li2022SeeTek,Chen2022PaLI}. How those features are aligned makes a big difference. 
In VDU, a majority of the tasks require \textit{local and layout-relative} understanding of the document. For example, in document VQA, semantic labeling or entity extraction, a model needs to make sense of text in-relation to where the text is placed in a document. E.g.: "1" when placed at the top-right/bottom-left of a document is to be interpreted as a page-number vs as a number when placed anywhere else. 

\begin{table}
\centering
	\scalebox{0.72}{
		\begin{tabular}{l|c|l|c|l|cH}
			Model & Year & Conf. & Arch. & Input Mod. & Vision Branch & Core Idea\\ 
	        \midrule
			LayoutLMv1 \cite{xu2020layoutlm} & 2020 & KDD & E & T + S & - & spatial emb.  \\
			DocFormerv1 \cite{appalaraju2021docformer} & 2021 & ICCV & E & T + V + S & Resnet50 & - \\
			LayoutLMv2 \cite{xu2020layoutlmv2} & 2021 & ACL & E & T + V + S & ResNeXt 101 & - \\
			SelfDoc \cite{Li2021SelfDoc} & 2021 & CVPR & E & - & - & - \\
			LayoutLMv3 \cite{huang2022layoutlmv3} & 2022 & ACM & E & T + V + S & Linear & - \\
			BROS \cite{bros2020hong} & 2022 & AAAI & E & T + S & - & -  \\ 
			XYLayoutLM \cite{Gu2022XYLayoutLM} & 2022 & CVPR & E & T + V + S & ResNeXt 101 & - \\
			FormNet \cite{Lee2022FormNet} & 2022 & ACL & E  & - & - & - \\
			ERNIE-Layout \cite{peng2022ernie} & 2022 & EMNLP & E & T + V + S & F-RCNN & - \\
			LiLT \cite{Wang2022LiLT} & 2022 & ACL & E & T + S & - & 3 on encoder \\
			XDoc \cite{Chen2022XDoc} & 2022 & EMNLP & E & T & - & - \\
			\midrule
			TILT \cite{powalski2021going} & 2021 & ICDAR & E + D & T + V + S & U-Net & 1 on decoder \\
			\midrule
			DocFormerv2 (ours) & 2023 & - & E + D & T + V + S & Linear & 2 task on enc, 1 on dec \\
		\end{tabular}
	}
	\caption{\textbf{VDU Related Work}: In this table, a summary of VDU prior art is presented  with their architecture (E: Encoder, D: Decoder), the input (T: text, V: vision, S: spatial features), the vision features branch and core idea behind the work.}
	\label{table:prior_art}
\end{table}

\begin{figure}
\centering
\includegraphics[width=0.5\textwidth]{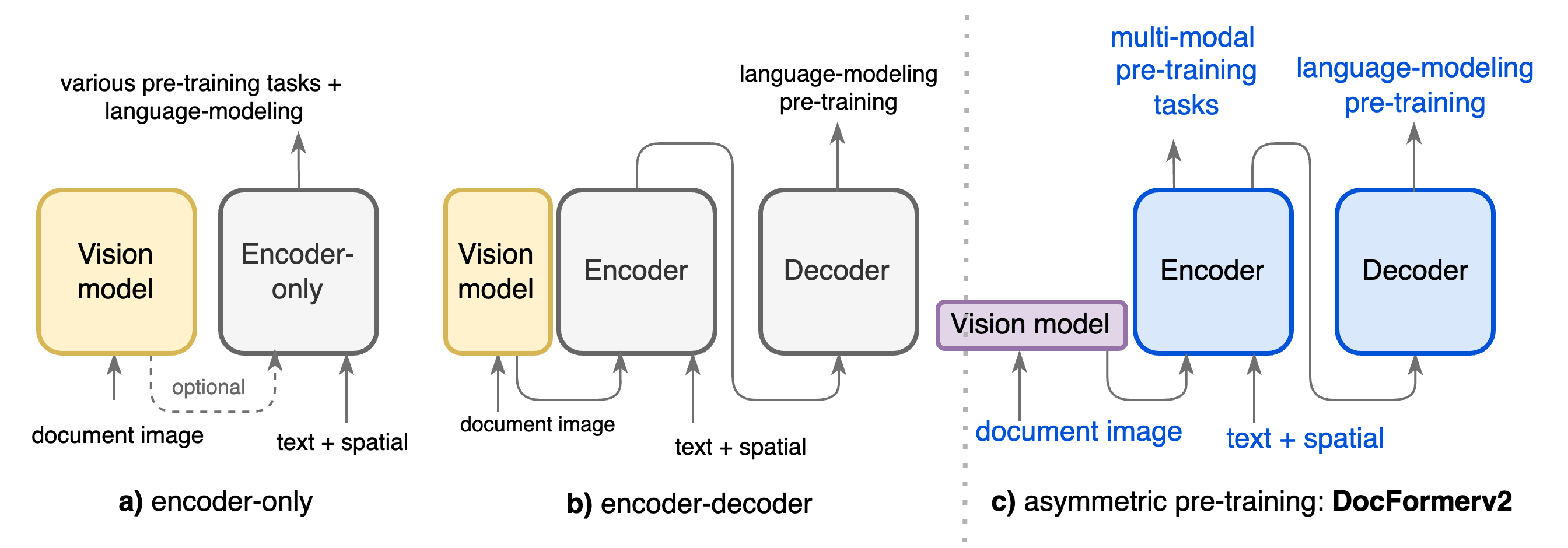}
\caption{\textbf{VDU Paradigms:} Broad state of Visual Document Understanding (VDU) approaches. In \textbf{a)} E-only LayoutLM \cite{xu2020layoutlm} and variants. \textbf{b)} E+D but only language-task TILT \cite{powalski2021going}. \textbf{c)} Ours }
\label{fig:prior_art}
\end{figure}



Based on this domain understanding of VDU and its challenges, we present \papertitle (\papertitleshortnospace) which is an encoder-decoder multi-modal transformer. 
In this work, we meticulously devise two novel unsupervised pre-training tasks with the objective of incorporating local semantic information of a document into the model. These pre-training tasks impart the ability to the model to accurately locate relevant information within the document. We also depart from VDU prior-art\cite{powalski2021going,Tang2022UnifyingUDOP} as we introduce a novel asymmetrical method of pre-training. i.e., multi-task pre-training on encoder (two tasks) and decoder (one task).
We propose two novel pre-training tasks on encoder with the intent to enrich the encoder with local semantic information. The tasks aid in by fusing and aligning multi-modal input and generating efficient representations for the decoder. 
We show that these pre-training tasks are necessary for effective VDU (see \S \ref{sec:experiments:ablation_experiments}). 
Furthermore, we demonstrate that a simplified linear visual layer is sufficient to encapsulate visual features, simplifying the architecture from previous VDU research \cite{xu2020layoutlmv2, Li2021SelfDoc, powalski2021going} which required specific visual encoders \cite{dosovitskiy2020image,Liu2021Swin, he_cvpr2016_resnet}.


Experimentally we demonstrate that \papertitle achieves state-of-the-art performance on  five VDU tasks. In addition, we demonstrate the versatility of \papertitle by utilizing its pre-trained model and fine-tuning it on text-VQA tasks from a completely different domain. Our approach yields superior performance on three distinct text-VQA datasets, surpassing comparable models and in some datasets much bigger models like GIT2 \cite{wang2022git}, PaLi \cite{Chen2022PaLI} and Flamingo \cite{Alayrac2022FlamingoAV}. 
Therefore, the primary contributions of this paper are as follows:


\begin{itemize}[leftmargin=*]
	\item Asymmetrical method of pre-training for VDU: Two novel tasks on the encoder which encourage local multi-modal feature collaboration (\textit{Token-to-Line} task and \textit{Token-to-Grid} task) and one on the decoder \S \ref{sec:approach:pretrain}. 
	\item Simplified Visual branch: \papertitle is end-to-end trainable and it does not rely on a pre-trained object detection network for visual features simplifying its architecture. On five varied downstream VDU tasks, \papertitle achieves state of the art results \S \ref{sec:experiments}. 
	\item We also show \papertitle versatility by fine-tuning it on a totally different domain - text-VQA datasets without changing the pre-training. \papertitle beats strong baselines and achieves state-of-the-art numbers on three text-VQA datasets amongst similar model sizes. Selectively, on Text-VQA it out-performs much larger models like PaLi-3B \textcolor{forestgreen}{+6.8\%}, PaLi-15B \textcolor{forestgreen}{+1.5\%} and Flamingo\cite{Alayrac2022FlamingoAV} \textcolor{forestgreen}{(+9.9\%)} (106x \papertitle size in the num. of parameters) by  absolute accuracy \S \ref{sec:experiments:textvqa}. 
\end{itemize}

\noindent Furthermore, we conducted comprehensive ablation experiments to demonstrate the advantages of our pre-training tasks, the model's resilience to input noise, and the efficacy of the simplified visual branch.

\begin{figure*}[t]
  \centering
  \includegraphics[width=0.7\linewidth]{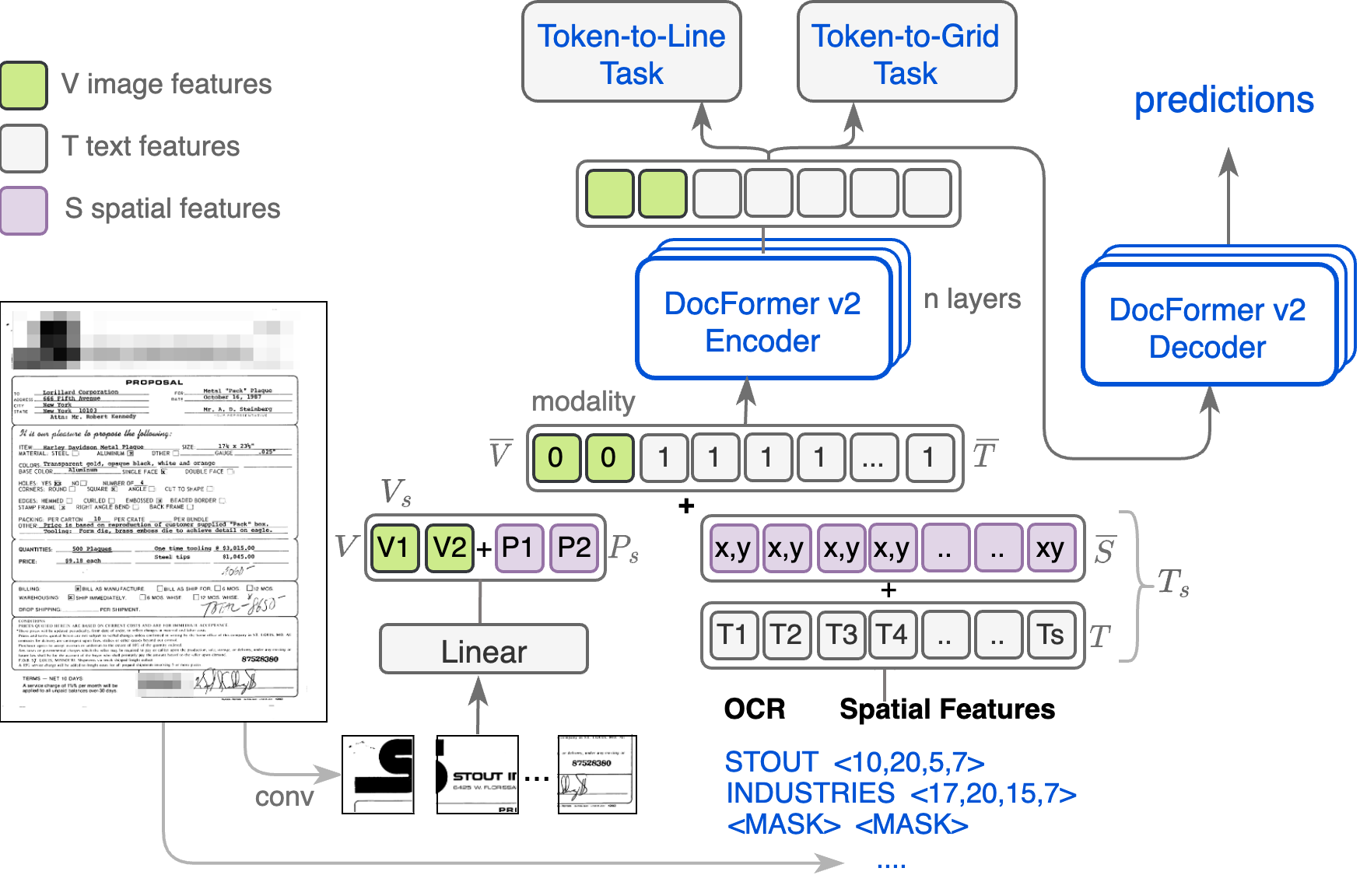}
  \caption{\textbf{\papertitle  Pre-train Architecture}. After pre-train, the two prediction heads (token-to-line and grid) on encoder are removed, rest of the architecture remains the same for down-stream tasks. Read section \ref{sec:approach:arch} for more details on $T_s$ and $V_s$. All components are end-to-end trainable. Best viewed in color. }
  \label{fig:docformerv2_pretrain_architecture}
\end{figure*}

\section{Related Work}
\label{sec:related_work}
VDU research has attracted considerable attention over the past few years \cite{Wang2022ABF,xu2020layoutlm,xu2020layoutlmv2,appalaraju2021docformer,Li2021SelfDoc,powalski2021going,Li2021MarkupLM,huang2022layoutlmv3,hong2020bros,Gu2022UnifiedUDOC,Gu2022XYLayoutLM,Lee2022FormNet,Wang2022LiLT,Chen2022XDoc,Tang2022UnifyingUDOP,Borchmann2021DUE,peng2022ernie,li2021structurallm}. Prominent published research papers in this area are catalogued in Table \ref{table:prior_art}. As can be observed, the research focus has been lopsided towards encoder-only models. While TILT \cite{powalski2021going} proposed a encoder-decoder transformer for VDU, they only train it on one pre-training task (masked language modeling) and also use a bulky visual CNN. Our approach \papertitle, not only simplifies the architecture by not using a separate visual module (CNN or Transformer based) and has multiple unsupervised pre-training tasks. 




\section{Approach}
\label{sec:approach}

\subsection{Architecture}
\label{sec:approach:arch}

\papertitle (\papertitleshort) is a multi-modal encoder-decoder transformer architecture (see fig. \ref{fig:docformerv2_pretrain_architecture}). Three variations of \papertitleshort are designed - small, base and large variants (see table \ref{table:dfv2_variants}).
\papertitleshort takes multi-modal inputs, the image of the document $I$, text $T$ extracted by an OCR model along with OCR bounding box co-ordinates as spatial features $\overline{S}$. \papertitleshort has a unified multi-modal encoder where the multi-modal features fuse and align with the help of novel pre-training tasks (see \S \ref{sec:approach:pretrain}). 

\label{sec:approach:arch:visual}
\noindent\textbf{Visual features:} \papertitleshort has a simplified visual branch contrary to most VDU prior-art (fig. \ref{fig:prior_art}). \papertitleshort consumes a flattened image sequence as visual input. Specifically, let $v \in \mathbb{R}^{3 \times h \times w}$ be the image of a document. A simple $V = linear( conv_{2\times2} ( v ) )$ is used to create an image embedding. The weights are randomly initialized for pre-training. As documents tend to have lots of white-space, the linear down-sampling layer gives an opportunity for the model to only keep relevant visual features. Based on our ablation experiments (see \S \ref{table:ablation:image_encoder}), this simple approach gives better results than using expensive image encoders such as Swin, ViT  \cite{Liu2021Swin, dosovitskiy2020image, Ronneberger2015UNet} or bulky object-detection networks like FRCNN variants \cite{Ren2015FasterFRCNN} as was used in VDU prior-art \cite{powalski2021going,appalaraju2021docformer,xu2021layoutlmv2}.
Since transformer layers are permutation-invariant, a learnable 2D-positional encoding $P_s$ is also computed. Finally, $V_s = V + P_s $

\label{sec:approach:arch:language}
\noindent\textbf{Language features:} Let $t$ be the predicted text extracted via an OCR model for a document image. \papertitleshort uses a sentence-piece sub-word tokenizer \cite{Kudo2018SentencePiece} to get tokens $t_{tok}$. A maximum sequence limit $s$ is applied during training and testing, so if the number of OCR tokens is greater than $s$, the rest is ignored. If the sequence length is less than $s$, the sequence is padded. The OCR tokens $t_{tok}$ are sent to a learnable embedding layer $W_t$ to create a text embedding $T= W_t(t_{tok})$.

\label{sec:approach:arch:spatial}
\noindent\textbf{Spatial features:} For each OCR word $t_i$, the OCR model predicts its bounding-box location in the normalized form $b_i = (x_1,y_1,x_3,y_3)$. This information is encoded using four learnable spatial embedding layers - $W_x$ for encoding a word horizontal spatial information $x_i$, $W_y$ for the vertical coordinate $y_i$, $W_h$ for word height $h_i$ and $W_w$ for the width $w_i$. The spatial features not only encode the location of a word in the document but also provides cues about a word's font-size and thereby its importance in a document (via $h_i$ and $w_i$). Specifically, spatial features  $\overline{S}= W_x(x_1, x_3) + W_y(y_1, y_3) + W_h(y_3-y_1) + W_w(x_3-x_1)$.  Finally, $T_s = T + \overline{S}$.


\label{sec:approach:arch:other}
\noindent\textbf{Other features:} $T_s$ and $V_s$ features are different modalities (fig. \ref{fig:docformerv2_pretrain_architecture}). As the model has no idea it is being fed multi-modal input, another learnable embedding $W_m$ is used to provide cues to the model about the multi-modal input. A modality-embedding $W_m$ learns nuances of different modalities, which generates $M_v$ embedding for visual modality and $M_t$ for text. Finally, $\overline{T} = T_s + M_t $ and $\overline{V} = V_s + M_v $. $\overline{T}$ and $\overline{V}$ are concatenated in the sequence dimension to form the input sequence to the \papertitleshort encoder.

\begin{table}
\centering
\scalebox{0.9}{
\begin{tabular}{c|c|c|c|c|c}
  model & dim & ff & \# attn. H & \# layers (E,D) & \# params  \\
  \midrule
  small & 512 & 2048 & 8 & 6,6 & 66M \\
  base & 768 & 3072 & 12 & 12,12 & 232M \\
  large & 1024 & 4096 & 16 & 24,24 & 750M \\
\end{tabular}
}
\caption{\textbf{\papertitle variants}: dim is embedding dimensionality. ff is output dim of feed-forward layer. E is encoder and D is decoder. attn. H is attention heads.
}
\label{table:dfv2_variants}
\vspace{-0.6cm}
\end{table}

\subsection{Unsupervised Document Pre-training}
\label{sec:approach:pretrain}

In \papertitle we follow the now well established practice of unsupervised pre-training followed by downstream task fine-tuning. Furthermore, with the intent of eliciting the maximum benefit from unsupervised pre-training, we designed the pre-training tasks as a close proxy for downstream tasks. We now describe the two novel pre-training tasks employed on the encoder and the language modeling task on decoder. All three tasks are performed at the same time and the final loss is a linear combination of all three losses for each iteration.

\begin{figure}[t]
\centering
\begin{minipage}[b]{0.45\linewidth}
\includegraphics[width=\linewidth]{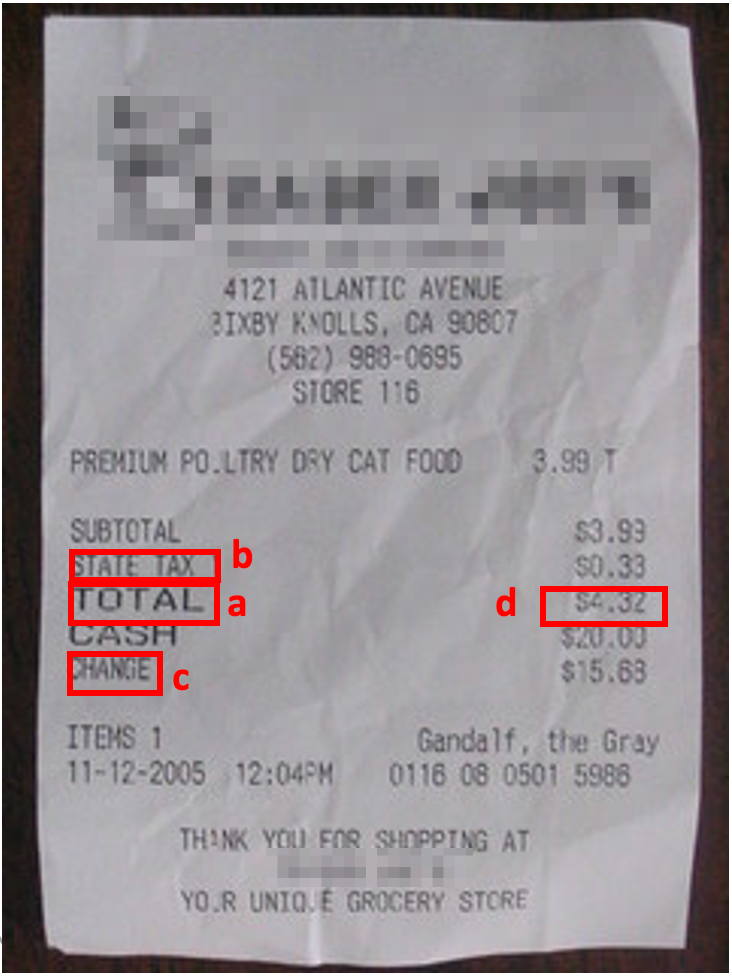}
\caption{\textbf{Token-to-Line}}
\label{fig:token_to_line}
\end{minipage}
\hspace{0.05\linewidth}
\begin{minipage}[b]{0.45\linewidth}
\includegraphics[width=\linewidth]{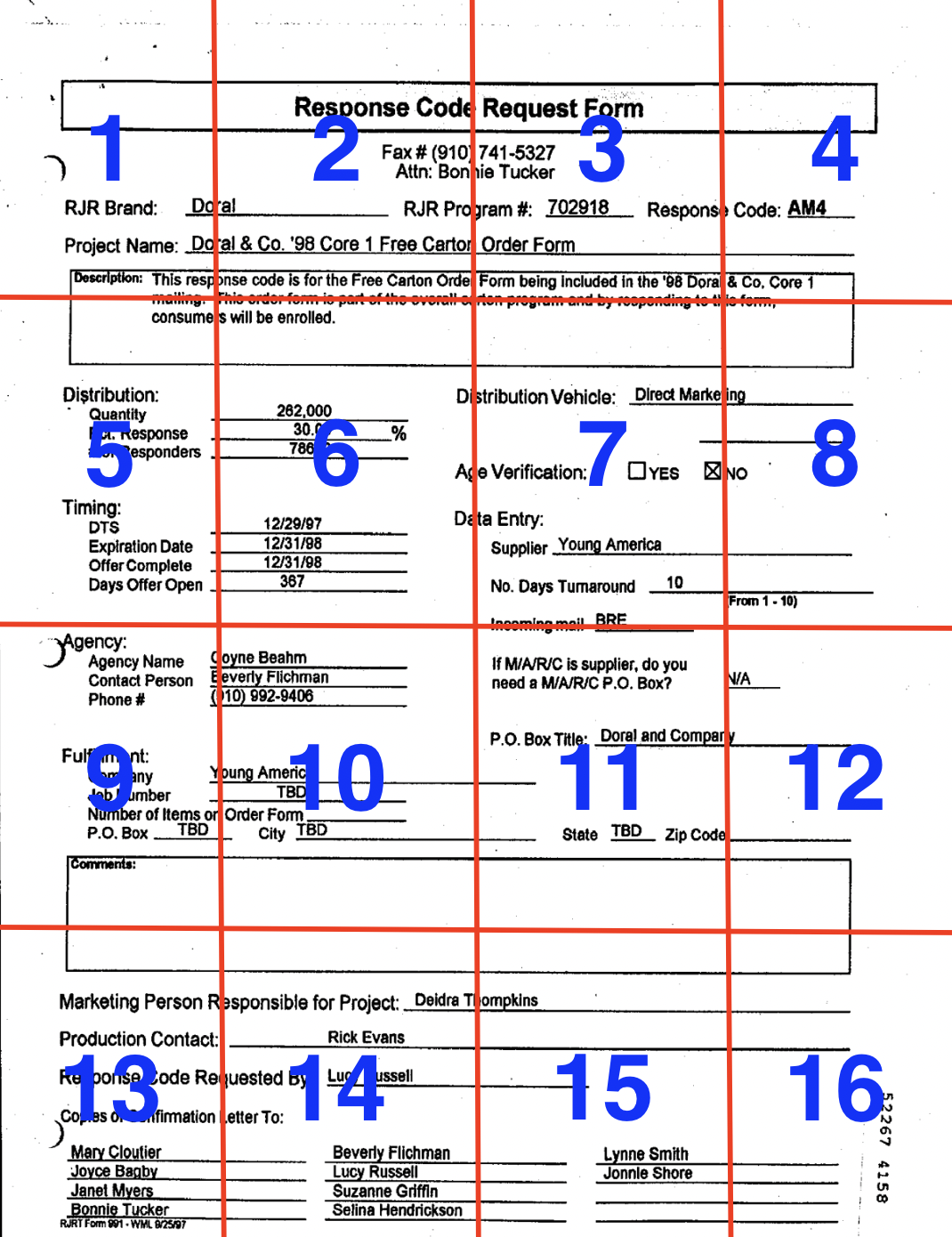}
\caption{\textbf{Token-to-Grid} 4x4}
\label{fig:token_to_grid}
\end{minipage}
\end{figure}

\label{sec:approach:pretrain:tok_to_line}
\noindent\textbf{Encoder Token-to-Line Task:} We share the intuition that for VDU tasks local feature semantic alignment is important. Most of the related information for key-value prediction in a form or VQA is either on the same line or adjacent lines of a document e.g., see fig. \ref{fig:token_to_line}, in order to predict the value for \verb|"TOTAL"| (box a), the model has to look in the same line (to its right - \verb|"$4.32"|  box d). We teach the model the relative position information between tokens. 
For implementation, we randomly pick two language tokens and ask the model to predict the number of lines between them. Furthermore, as a document could have an arbitrary number of lines of text, the task is quantized. i.e., there are only three labels: \verb|{0, 1, 2}|. All token pairs that are more than 2 lines apart are labelled as 2 because distant tokens are not likely related and the model should learn to ignore them. Assume that $a, b, c, d$ (fig. \ref{fig:token_to_line}) are lines. Let $F$ be the \papertitleshort encoder head function trying to predict a label for this task. then: 

\vspace{-3mm}
\begin{equation}
F(a, d) = 0 ; \
F(a, b) = 1 ; \
F(b, c) = 2
\end{equation}

\noindent Based on the ablation (table \ref{table:ablation:pretrain_tasks}), this task gives \textcolor{forestgreen}{+2.2\%}  benefit on DocVQA task. The loss for this task is tracked as $L_{tol}$.

\label{sec:approach:pretrain:tok_to_grid}
\noindent\textbf{Encoder Token-to-Grid Task:} Different semantic information is concentrated in different regions of the document. For example, a) In a financial document, the top block contains the header, the middle block contains information to be filled and the bottom block typically contains footer elements/instructions. b) Page numbers are typically at the top or the bottom. c) In a receipt/invoice the company name is typically at the top. The content of a document is presented according to a visual layout and structure that must be taken into account for accurate understanding. Based on this intuition, this task pairs language semantics with the location (visual, spatial or both) in a document. Specifically, the document is virtually divided into a m x n grid. Each OCR token is assigned a grid number and \papertitleshort is tasked with predicting the grid number for each token. For each OCR token $t_i$, its top-left location $(x_1, y_1)$ is used to determine its grid-number $g_i$. Grids are in raster-scan order, so if a particular token falls on the boundary of multiple grids, the scan-order is used to disambiguate. If a token falls on the boundary of normalized image co-ordinates, they are ignored for prediction. See fig. \ref{fig:token_to_grid} for viz. Specifically, we have: 

$g_i = \lfloor\frac{x_1}{\Delta_x}\rfloor + \lfloor\frac{y_1}{\Delta_y}\rfloor\cdot m,$

\noindent where $\Delta_x$ and $\Delta_y$ are the widths and heights of each grid, respectively, and $m$ is the number of grids in a row. The loss $L_{tog}$.



\noindent \textbf{Decoder Language Modeling:} Since VDU predictions are in the language domain, language understanding forms an important component of \papertitleshort pre-training. We do the denoising masked language modeling popularized by T5 \cite{Raffel2019ExploringT5}.  During pre-training, not only are the input tokens randomly \verb|MASKED| it's spatial features (mentioned in \S \ref{sec:approach:arch:spatial}) are also masked. Masking the spatial features $\overline{S}$ for the masked tokens makes the grid prediction and line prediction hard because the model does not have 2D-position information of the masked tokens. It has to infer from other available context. The loss for this operation is denoted $L_{dlm}$.

\noindent \textbf{Final pre-training loss:} The final loss is a linear combination of all three pre-training losses i.e., $L_{final} = k * L_{tol} + l * L_{tog} + m * L_{dlm}$, where $k, l, m$ are empirically determined. 

\noindent \textbf{Downstream Tasks:} Once pre-training is done, we remove the token-to-line and token-to-grid linear prediction heads. The rest of the pre-trained model is fine-tuned on the respective downstream train data.


\section{Experiments}
\label{sec:experiments}

\noindent \textbf{Implementation details}: Following prior-art \cite{appalaraju2021docformer,powalski2021going,biten2022latr,xu2020layoutlm,xu2021layoutlmv2,huang2022layoutlmv3} we use the Industrial Document Library (IDL)\footnote{https://www.industrydocuments.ucsf.edu/} dataset for pre-training. The IDL is a collection of industry documents hosted by UCSF. It hosts millions of documents publicly disclosed from various industries like tobacco, drug, food etc. The data from the website amounts to about 13M documents, translating to about 70M pages of various document images. We further extracted OCR for each document. 
Data was cleaned and about 6M documents were pruned, the resulting 64M document images and OCR-text (with spatial co-ordinates) is used for unsupervised pre-training. The data distribution for IDL 64M is presented in supplemental section.

\noindent \textbf{Downstream experiments:} The model is fine-tuned on the provided training set and numbers are reported on the corresponding validation/test set. No dataset specific hyper-parameter tuning is done. This is an advantage of our approach and we believe that the numbers may be higher if dataset specific fine-tuning is done. Details about fine-tuning datasets are in the supplemental section. We used Pytorch \cite{paszke2019pytorch} and the Huggingface library \cite{wolf2019huggingface}.

\noindent \textbf{Evaluation Metrics:} A dataset specific evaluation metric is adopted. For DocVQA\cite{mathew2020docvqa}, InfoVQA\cite{mathew2022infographicvqa}, ST-VQA\cite{biten2019scene}, Average Normalized Levenshtein Similarity (ANLS) \cite{biten2019icdar} is used. ANLS measures the similarity between the predicted results and ground truth and ranges from (0,100). For FUNSD\cite{Jaume2019FUNSDAD}, CORD\cite{park2019cord} F1-score is used. For TextVQA \cite{singh2019towards} and OCR-VQA\cite{mishra2019ocr}  accuracy is used. In all metrics, higher the better.

\subsection{Table VQA}

\textbf{TabFact} \cite{Chen2019TabFactAL, Borchmann2021DUE}: This dataset studies table understanding and fact verification with semi-structured evidence over 
tables collected from Wikipedia. Entailed and refuted statements corresponding to a single row or cell were prepared by the authors of TabFact \cite{Chen2019TabFactAL}. This task poses challenges due to the complex linguistic and spatial reasoning involved. In Table \ref{table:tablevqa}, we can see that \papertitle out-performs prior art by a large margin \textcolor{forestgreen}{(+4.3\%)}.

\begin{table}[htbp]
\centering
\scalebox{0.81}{
\begin{tabular}{l|c|cH|c}
  Model & \thead{pre-train\\data} & \thead{\#param} & \thead{\textbf{WikiTable} \\Acc. (\%)} & \thead{\textbf{TabFact} \\Acc. (\%)} \\
  \midrule
  \multicolumn{5}{l}{\textit{methods based on only text / (text + spatial) features:}} \\ 
  \midrule
  T5$_{\text{large}}$\cite{Raffel2019ExploringT5} & - & 750M & 33.3 & 58.9 \\
  T5$_{\text{large}}$+U \cite{Borchmann2021DUE} & - & 750M & 38.1 & 76.0 \\
  T5$_{\text{large}}$+2D \cite{Borchmann2021DUE} & - & 750M & 30.8 & 58.0 \\
  T5$_{\text{large}}$+2D+U \cite{Borchmann2021DUE} & - & 750M & 43.3 & 78.6 \\
  \midrule
  \multicolumn{5}{l}{\textit{methods based on image + text + spatial features:}} \\ 
  \midrule
  LayoutLMv3$_{\text{large}}$ \cite{huang2022layoutlmv3} & 11M & 368M & 45.7 & 78.1 \\
  UDOP \cite{Tang2022UnifyingUDOP} & 11M & 794M & \textbf{47.2} & \underline{78.9} \\
  \midrule
  \papertitlenospace$_{\text{large}}$ & 64M & 750M & \underline{47.1} & \textbf{83.2}
  \textcolor{forestgreen}{(+4.3\%)}  \\
\end{tabular}
}
\caption{\textbf{Comparison on Table VQA Datasets}\cite{Borchmann2021DUE}: Our work, \papertitle outperforms the previous state of the art. 
}
\label{table:tablevqa}
\vspace{-0.4cm}
\end{table}

\subsection{Document VQA}

DocVQA \cite{mathew2020docvqa} and InfographicsVQA \cite{mathew2022infographicvqa} are datasets for the document VQA task.
DocVQA \cite{mathew2020docvqa} focuses on VQA for real-world industry documents and requires that the model understand images, texts, tables, forms, \etc.
InfographicsVQA \cite{mathew2022infographicvqa} focuses on VQA for infographics and requires that the model understand plots/graphs, texts, layout, figures. A model needs to reason multi-modally to generate an answer for this data. Please see the supplemental for data statistics and samples.

\begin{table}[htbp]
\centering
\scalebox{0.75}{
\begin{tabular}{l|c|c|c|c}
  Model & \thead{pre-train\\data} & \thead{\#param} & \thead{\textbf{DocVQA} \\test ANLS (\%)} & \thead{\textbf{InfoVQA} \\test ANLS (\%)} \\
  \midrule
  \multicolumn{5}{l}{\textit{methods based on only image:}} \\
  \midrule
  Donut$_{\text{base}}$ \cite{Kim2022DonutECCV} & 13M & 143M & 67.5 & 11.5\\
  Pix2Struct$_{\text{large}}$ \cite{Lee2022Pix2Struct} & 80M & 1.3B & 76.6 & 40.0 \\
  \midrule
  \multicolumn{5}{l}{\textit{methods based on only text / (text + spatial) features:}} \\ 
  \midrule
  T5$_{\text{large}}$\cite{Raffel2019ExploringT5} & - & 750M & 70.4 & 36.7 \\
  T5$_{\text{large}}$+U \cite{Borchmann2021DUE} & - & 750M & 76.3 & 37.1 \\
  T5$_{\text{large}}$+2D \cite{Borchmann2021DUE} & - & 750M & 69.8 & 39.2 \\
  T5$_{\text{large}}$+2D+U \cite{Borchmann2021DUE} & - & 750M & 81.0 & 46.1 \\
  \midrule
  \multicolumn{5}{l}{\textit{methods based on image + text + spatial features:}} \\ 
  \midrule
  LayoutLMv3$_{\text{large}}$ \cite{huang2022layoutlmv3} & 11M & 368M & 83.4 & 45.1 \\
  UDOP \cite{Tang2022UnifyingUDOP} & 11M & 794M & 84.7 & \underline{47.4} \\
  LayoutLMv2$_{\text{large}}^{\dagger}$ \cite{xu2020layoutlmv2} & 11M & 426M & 86.7 & - \\
  TILT$_{\text{large}}^{\dagger}$ \cite{powalski2021going} & 1.1M & 780M & \underline{87.05} & - \\
  \midrule
  \papertitlenospace$_{\text{large}}$ & 64M & 750M & 87.2 & - \\
  \papertitlenospace$_{\text{large}}^{\dagger}$ & 64M & 750M & \textbf{87.84} \textcolor{forestgreen}{(+0.79\%)} & \textbf{48.8} \textcolor{forestgreen}{(+1.4\%)} \\
\end{tabular}
}
\caption{\textbf{Comparison on Document VQA Datasets}\cite{mathew2020docvqa,mathew2022infographicvqa}: Our work, \papertitle outperforms the previous state of the art. $^{\dagger}$ indicates training with extra document VQA data.
}
\label{table:docvqa}
\vspace{-0.2cm}
\end{table}

Following common practice \cite{Borchmann2021DUE,powalski2021going,xu2020layoutlmv2}, we train \papertitle on the combination of the training and validation sets and do evaluation on the test set for each dataset.
In addition, we also follow \cite{powalski2021going,xu2020layoutlmv2} to train \papertitle on an extra document VQA dataset with 850k question-answer pairs and then fine-tune on DocVQA/InfographicsVQA for higher accuracy.

\papertitle outperforms (Table~\ref{table:docvqa})  the previous state of the art for document VQA even without using any extra document VQA pre-training data.
After pre-training on the extra data, \papertitle surpasses the previous state of the art by 0.79\% on DocVQA and 1.4\% on InfographicsVQA, which confirms the effectiveness of our approach.

\begin{table}
	\centering
	\scalebox{0.75}{
		\begin{tabular}{l|c|c|c|c}
			Model & \#param & Precision & Recall & F1 \\ 
			\Xhline{2.5\arrayrulewidth}
			\multicolumn{5}{l}{\textit{methods based on only image:}} \\
			\midrule
			Dessurt$_{\text{base}}$  \cite{Davis2022EndtoendDesseurt} & 127M & - & - & 65.0  \\
			\midrule
			\multicolumn{5}{l}{\textit{methods based on only text / (text + spatial) features:}} \\ \midrule
			BERT$_{\text{base}}$ \cite{devlin2018bert} & 109M & 54.69 & 61.71 & 60.26  \\
			RoBERTa$_{\text{base}}$ \cite{liu2019roberta} & 125M & 63.49 & 69.75 & 66.48 \\
			UniLMv2$_{\text{base}}$ \cite{bao2020unilmv2} & 125M & 63.49 & 69.75 & 66.48  \\ 
			LayoutLMv1$_{\text{base}}$ \cite{xu2020layoutlm} & 113M & 76.12 & 81.55 & 78.66  \\
			BROS$_{\text{base}}$ \cite{bros2020hong} & 139M & 80.56 & 81.88 & 81.21  \\ 
			BERT$_{\text{large}}$ \cite{devlin2018bert} & 340M & 61.13 & 70.85 & 65.63 \\
			RoBERTa$_{\text{large}}$ \cite{liu2019roberta} & 355M & 67.80 & 73.91 & 70.72 \\
			UniLMv2$_{\text{large}}$  \cite{bao2020unilmv2} & 355M & 67.80 & 73.91 & 70.72  \\ 
			LayoutLMv1$_{\text{large}}$ \cite{xu2020layoutlm} & 343M & 75.36 & 80.61 & 77.89  \\
			StructuralLM$_{\text{large}}$ \cite{li2021structurallm}&355M&83.52&86.81&85.14\\
			FormNet \cite{Lee2022FormNet}&217M&85.21&84.18&84.69\\
			\midrule
			\multicolumn{5}{l}{\textit{methods based on image + text + spatial features:}} \\ \midrule
			LayoutLMv1$_{\text{base}}$  \cite{xu2020layoutlm} & 160M & 76.77 & 81.95 & 79.27  \\
			LayoutLMv2$_{\text{base}}$ \cite{xu2020layoutlmv2}  & 200M & 80.29 & 85.39 & 82.76  \\ 
			LayoutLMv2$_{\text{large}}$ \cite{xu2020layoutlmv2}  & 426M & 83.24 & 85.19 &  84.20   \\ 
			DocFormer$_{\text{base}}$ \cite{appalaraju2021docformer} & 183M & 80.76 & 86.09 & 83.34 \\ 
			DocFormer$_{\text{large}}$ \cite{appalaraju2021docformer} & 536M & 82.29 & 86.94 &  84.55  \\
			SelfDoc \cite{Li2021SelfDoc}&-&-&-&83.36\\
			UDoc \cite{Gu2022UnifiedUDOC}&272M&-&-&\underline{87.93}\\
			StrucTexT \cite{li2021structext}\ding{66}&107M& 85.68 &80.97 &83.09\\
			 LayoutLMv3$_{\text{base}}$ \cite{huang2022layoutlmv3}\ding{101}&133M&77.39&81.65&79.46 \\
			 LayoutLMv3$_{\text{large}}$ \cite{huang2022layoutlmv3}\ding{101}&368M&81.35&83.75&82.53 \\
			\textcolor{codegray}{LayoutLMv3$_{\text{base}}$} \cite{huang2022layoutlmv3}\ding{109}&133M&89.55&91.65&\textcolor{codegray}{90.29}\\
			\textcolor{codegray}{LayoutLMv3$_{\text{large}}$} \cite{huang2022layoutlmv3}\ding{109}&368M&92.19&92.10&\textcolor{codegray}{92.08}\\
			\textcolor{codegray}{UDOP \cite{Tang2022UnifyingUDOP}}\ding{109} &794M&-&-&\textcolor{codegray}{91.62}   \\
			\midrule
			\papertitlenospace$_{\text{base}}$  &232M&89.15&87.6&88.37  \\
		    \papertitlenospace$_{\text{large}}$&750M&89.88&  {87.92}&{\bf88.89} \\
			
		\end{tabular}
	}
	\caption{\textbf{FUNSD comparison}: \papertitle does better than  models its size and compares well with even larger models. \ding{66} does not use standard train/test split, and the results are not directly compared with others. \ding{109} use OCR lines (not word box) as 2D position for words, and use entity boxes as 2D position for each word during finetuning and test, and thus the results are not directly comparable. \ding{101} are results by using the word boxes as 2D position for each word as other competitors do.}
	\label{table:funsd}
	\vspace{-2.5ex}
\end{table}


\begin{table}[!h]
	\scalebox{0.8}{
		\begin{tabular}{l|c|c|c|c}
			Model & \#param & Precision & Recall & F1 \\ 
			\Xhline{2.5\arrayrulewidth}
			\multicolumn{5}{l}{\textit{methods based on only text / (text + spatial) features:}} \\ \midrule
			BERT$_{\text{base}}$ \cite{devlin2018bert} & 109M & 88.33 & 91.07 & 89.68  \\
			UniLMv2$_{\text{base}}$  \cite{bao2020unilmv2} & 125M & 89.87 & 91.98 & 90.92  \\ 
			SPADE \cite{hwang2020spatial} & - & - & - & 91.50 \\
			LayoutLMv1$_{\text{base}}$ \cite{xu2020layoutlm} & 113M & 94.37 & 95.08 & 94.72  \\
			BROS$_{\text{base}}$ \cite{bros2020hong}   & 139M & 95.58 & 95.14 & 95.36  \\ 
			BERT$_{\text{large}}$ \cite{devlin2018bert} & 340M & 88.86 & 91.68 & 90.25 \\
			RoBERTa$_{\text{large}}$ \cite{liu2019roberta}&355M&-&-&93.80\\
			UniLMv2$_{\text{large}}$  \cite{bao2020unilmv2} & 355M & 91.23 & 92.89 & 92.05 \\ 
			LayoutLMv1$_{\text{large}}$ \cite{xu2020layoutlm} & 343M & 94.32 & 95.54 & 94.93 \\
			FormNet \cite{Lee2022FormNet}&345M&98.02&96.55& \underline{97.28}\\
			\midrule
			\multicolumn{5}{l}{\textit{methods based on image + text + spatial features:}} \\ \midrule
			LayoutLMv2$_{\text{base}}$ \cite{xu2020layoutlmv2}  & 200M & 94.53 & 95.39 & 94.95  \\ 
			LayoutLMv2$_{\text{large}}$ \cite{xu2020layoutlmv2} & 426M & 95.65 & 96.37 & 96.01 \\
			TILT$_{\text{base}}$ \cite{powalski2021going}\ding{109} & 230M & - & - & 95.11 \\
			TILT$_{\text{large}}$ \cite{powalski2021going}\ding{109} & 780M & - & - & 96.33 \\
			DocFormer$_{\text{base}}$\cite{appalaraju2021docformer} & 183M & 96.52 & 96.14 &  96.33  \\
			DocFormer$_{\text{large}}$ \cite{appalaraju2021docformer} & 536M & 97.25 & 96.74 &  96.99  \\
			UDoc \cite{Gu2022UnifiedUDOC}&272M&-&-&96.86 \\
		 LayoutLMv3$_{\text{base}}$ \cite{huang2022layoutlmv3}\ding{101}&133M&92.92&94.31&93.61 \\
			 LayoutLMv3$_{\text{large}}$ \cite{huang2022layoutlmv3}\ding{101}&368M&96.78&96.78&96.78\\
			\textcolor{codegray}{LayoutLMv3$_{\text{base}}$} \cite{huang2022layoutlmv3}\ding{109}&133M&-&-&\textcolor{codegray}{ 96.56}\\
			\textcolor{codegray}{LayoutLMv3$_{\text{large}}$} \cite{huang2022layoutlmv3}\ding{109}&368M&-&-&\textcolor{codegray}{97.46}\\
			\textcolor{codegray}{UDOP \cite{Tang2022UnifyingUDOP}}\ding{109} &794M&-&-&\textcolor{codegray}{97.58}   \\
			\midrule
			\papertitlenospace$_{\text{base}}$ & 232M & 97.51 & 96.10  &96.80\\
			\papertitlenospace$_{\text{large}}$ & 750M & 97.71 & 97.70 & \textbf{97.70} \\
		\end{tabular}
	}
	\caption{\textbf{CORD dataset} \cite{park2019cord} \textbf{comparison}. We present entity-level Precision, Recall, F1 on test set. \ding{109} use OCR lines (not word box) as 2D position for words, and use entity boxes as 2D position for each word during finetuning and testing, and thus the results are not directly comparable. \ding{101} are results by using the word boxes as 2D position for each word as the other competitors do.}
	\label{table:cord}
\end{table}

\subsection{Sequence Labeling Task}
We study the performance of DocFormerv2 on the semantic entity-labeling task (i.e., group tokens which belong to the same class). We test the model on FUNSD dataset \cite{Jaume2019FUNSDAD}, which is a forms dataset containing 199 noisy documents (149 images for train, 50 images for test). There are four classes: {\em question, answer, header}, and {\em other}. We measure entity-level performance using F1 score (Table \ref{table:funsd}). The input sequence to Docformerv2 includes individual texts as prompts and all document texts as context, and the decoder sequence contains the entity texts and predicted labels. Docformerv2 achieves 88.89\% F1 score (Table \ref{table:funsd}), and outperforms the existing methods without using entity box priors in pretraining and finetuning (grayed models in the table). 

\subsection{Entity Extraction Task}
We evaluate DocFormerv2 for the entity extraction task on the CORD dataset.
CORD \cite{park2019cord} consists of 1000 receipts (800/100/100 images for train/val/test). It defines 30 fine-grained fields under 4 coarse-grained categories. To extract all entities, in the input sequence, we add a question of ``{\em What are entities of $<$CLASS$>$?}" in front of all text context tokens. The output of the decoder includes all entities which are separated by a separator token. Following the standard evaluation metric for entity extraction, we measure entity-level performance using F1 score.
Docformerv2 (Table \ref{table:cord}) achieves 97.7\% F1 score, and outperforms existing methods. Docformerv2 enables multiple entities decoding in an auto-regressive way which shows that the model is able to learn both intra-entity and inter-entity structures.



\begin{table}
\centering
\scalebox{0.81}{
\begin{tabular}{l|c|c|c|c}
  Model & \thead{pre-train\\data} & \thead{\#param} & \thead{Val\\Acc. (\%)} & \thead{Test\\ Acc. (\%)} \\
  \midrule
  Blk+CNN+W2V & - & - & - & 48.3 \\
  M4C \cite{hu2020iterative}& - & 200M & 63.5 & 63.9 \\
  LaAP \cite{han2020finding} & - & - & 63.8 & 64.1 \\
  LaTr$_{\text{base}}$ \cite{biten2022latr}& 64M & 311 & 67.5 & 67.9 \\
  \midrule
  GIT$_{\text{base}}$ & 10M & 129M & 57.3 & 57.5 \\
  GIT$_{\text{large}}$ & 20M & 347M & 62.4 & 62.9 \\
  GIT & 800M & 681M & 67.8 & \underline{68.1} \\
  \textcolor{codegray}{GIT2}\ding{66} \cite{wang2022git} & \textcolor{codegray}{12.9B} & \textcolor{codegray}{5.1B} & - & \textcolor{codegray}{70.3} \\
  \midrule
  \papertitlenospace$_{\text{base}}$ & 64M & 232M & 69.7 & 70.3 \\
  \papertitlenospace$_{\text{large}}$ & 64M & 750M & 71.1 & \textbf{71.5} \textcolor{forestgreen}{(+3.4\%)}  \\
\end{tabular}
}
\caption{\textbf{Comparison on OCR-VQA}\cite{mishra2019ocr}: \papertitle is better than the previous SOTA by \textcolor{forestgreen}{(+3.4\%)}. \textbf{Bold} indicates best and \underline{underline} indicates the previous state of the art.  GIT2 \ding{66}: uses extra VQA data (aggregation of 8 VQA datasets).
}
\label{table:ocrvqa}
\vspace{-0.6cm}
\end{table}


\begin{table}
\centering
\scalebox{0.81}{
\begin{tabular}{l|l|c|c|c}
  Model & \thead{pre-train\\data} & \thead{\#param} & \thead{Val\\Acc. (\%)} & \thead{Test\\ Acc. (\%)} \\
  \midrule
  M4C \cite{hu2020iterative} & - & 200M &  47.8 & - \\
  LaAP \cite{han2020finding} & - & - & 41.0 & 41.4 \\
  SA-M4C \cite{kant2020spatially} & - & 200M & 45.4 & 44.6 \\
  SMA & - & - & 44.6 & 45.5 \\
  SceneGate \cite{Luo2022SceneGATESB} & - & - & 42.4 & 44.0 \\
  SC-Net \cite{Fang2022TowardsEF} & - & - & 44.8 & 45.7 \\
  LOGOS \cite{lu2021localize} & - & - & 51.5 & 51.1 \\
  TAP + TAG \cite{wang2022tag} & - & - &  53.6 & 53.7 \\
  TAP \cite{yang2021tap} & - & 200M & 54.7 & 54.0 \\
  TAP Two-Stage \cite{Li2022TwoStageTAP} & - & 200M & 55.9 & 55.3 \\
  \textcolor{codegray}{Flamingo} \cite{Alayrac2022FlamingoAV} & \textcolor{codegray}{2.3B}\ding{72} & \textcolor{codegray}{80B} & \textcolor{codegray}{57.1} & \textcolor{codegray}{54.1} \\
  PreSTU \cite{Kil2022PreSTUPF} & 13M & 237M & - & 56.3 \\
  \midrule
  LaTr$_{\text{base}}$ \cite{biten2022latr} & 64M & 311M & 58.0 & 58.9 \\
  LaTr$_{\text{base}}^{\dagger}$\cite{biten2022latr} & 64M & 311 & 59.5 & 59.6 \\
  LaTr$_{\text{large}}^{\dagger}$\cite{biten2022latr} & 64M & 856M & 61.0 & \underline{61.6} \\
  \midrule
  GIT$_{\text{base}}$ & 10M\ding{109} & 129M & 18.8 & - \\
  GIT$_{\text{large}}$ & 20M\ding{109} & 347M & 37.5 & - \\
  GIT & 800M\ding{109} & 681M & 59.9 & 59.8 \\
  \textcolor{codegray}{GIT2}\ding{66} \cite{wang2022git} & \textcolor{codegray}{12.9B\ding{109}} & \textcolor{codegray}{5.1B} & \textcolor{codegray}{68.4} & \textcolor{codegray}{67.3} \\  
  \midrule
  \textcolor{codegray}{PaLi-3B} \cite{Chen2022PaLI} & \textcolor{codegray}{1.6B}\ding{109} & \textcolor{codegray}{3B} & \textcolor{codegray}{58.8} & - \\
  \textcolor{codegray}{PaLi-15B} \cite{Chen2022PaLI} & \textcolor{codegray}{1.6B}\ding{109} & \textcolor{codegray}{15B} & \textcolor{codegray}{64.1} & - \\
  \textcolor{codegray}{PaLi-17B} \cite{Chen2022PaLI} & \textcolor{codegray}{1.6B\ding{109}} & \textcolor{codegray}{17B} & \textcolor{codegray}{70.5} & \textcolor{codegray}{73.1} \\  
  \midrule
  \papertitlenospace$_{\text{base}}^{\dagger}$ & 64M & 232M & 61.6 & 60.0 \\
  \papertitlenospace$_{\text{large}}^{\dagger}$ & 64M & 750M &  \textbf{65.6} & \textbf{64.0} \textcolor{forestgreen}{(+2.4\%)} \\
  
\end{tabular}
}
\caption{\textbf{Comparison on TextVQA}\cite{singh2019towards}: $^{\dagger}$ indicates the model used the combination of ST-VQA and TextVQA training sets to train the model. GIT2 \ding{66}: extra data used (aggregation of 8 VQA datasets) \ding{72}: video-text data. \ding{109}: proprietary image-text data. \textcolor{codegray}{Grey rows} shows models which are much bigger (\# parameters $\ge$ 3x \papertitleshort$_{\text{large}}$ parameters) and use large amounts of external data. \papertitleshort$_{\text{large}}$ still outperforms Flamingo \textcolor{forestgreen}{(+9.9\%)}, Pali-3B \textcolor{forestgreen}{(+6.8\%)} and Pali-15B \textcolor{forestgreen}{(+1.5\%)} models. 
}
\label{table:textvqa}
\vspace{-0.3cm}
\end{table}


\begin{table}[t]
\centering
\scalebox{0.8}{
\begin{tabular}{l|c|c|c|c}
  Model & \thead{pre-train\\data} & \#param &  \thead{Val\\ ANLS (\%)} & \thead{Test\\ ANLS (\%)} \\
  \midrule
  M4C \cite{hu2020iterative} & - & 200M & 47.2 & 46.2 \\
  LaAP \cite{han2020finding} & - & - & 49.7 & 48.5 \\
  SA-M4C \cite{kant2020spatially} & - & 200M & 51.2 & 50.4 \\
  SceneGate \cite{Luo2022SceneGATESB} & - & - & 52.5 & 51.6 \\
  LOGOS \cite{lu2021localize} & - & - & 58.1 & 57.9 \\
  TAP \cite{yang2021tap} & & 200M & 59.8 & 59.7 \\
  TAP + TAG \cite{wang2022tag} & - & - & 62.0 & 60.2 \\
  PreSTU \cite{Kil2022PreSTUPF} & 13M & 237M & - & 65.5 \\
  \midrule
  LaTr$_{\text{base}}$ \cite{biten2022latr} & 64M & 311M & 67.5 & 66.8 \\
  LaTr$_{\text{base}}^{\dagger}$\cite{biten2022latr} & 64M & 311M & 68.3 & 68.4 \\
  LaTr$_{\text{large}}^{\dagger}$\cite{biten2022latr} & 64M & 856M &70.2 & \underline{69.6} \\
  \midrule
  GIT$_{\text{base}}$ & 10M\ding{109} & 129M & 20.7 & - \\
  GIT$_{\text{large}}$ & 20M\ding{109} & 347M & 44.6 & - \\
  GIT & 800M\ding{109} & 681M & 69.1 & \underline{69.6} \\
  \midrule
  \papertitlenospace$_{\text{base}}^{\dagger}$ & 64M & 232M & 70.1 & 68.4 \\
  \papertitlenospace$_{\text{large}}^{\dagger}$ & 64M & 750M & \textbf{72.9} & \textbf{71.8} \textcolor{forestgreen}{(+2.2\%)} \\
\end{tabular}
}
\caption{\textbf{Comparison on ST-VQA}\cite{biten2019scene}: $^{\dagger}$ indicates the combination of the ST-VQA and TextVQA training sets is used.
}
\label{table:stvqa}
\vspace{-0.4cm}
\end{table}

\subsection{Generalization Experiments - Scene-Text VQA}
\label{sec:experiments:textvqa}

In this section, we show the strength of \papertitle on a different task - Text-VQA. Unlike document understanding which focuses on document images, the Text-VQA task answers questions for natural images with scene text. 
We fine-tune our \textit{document} pre-trained models on three Text-VQA datasets. We emphasize that no image-text pre-training was performed on \papertitlenospace, it was merely fine-tuned on the respective Text-VQA training dataset.
Three popular Text-VQA datasets are used - OCR-VQA \cite{mishra2019ocr}, TextVQA \cite{singh2019towards} and ST-VQA \cite{biten2019scene}, each with strong baselines from the vision-language community (as is standard practice by Text-VQA we mean any scene text VQA dataset while TextVQA refers to a specific dataset). Please see the supplemental for a dataset breakdown.
For OCR-VQA, we fine-tune our models on the training set and do evaluation on the validation and test sets. For TextVQA and ST-VQA, following the previous state-of-the-art methods \cite{biten2022latr,yang2021tap}, we fine-tune our models on the combination of the TextVQA and ST-VQA training sets and do evaluation on the validation and test sets of each dataset.
Tables \ref{table:ocrvqa}, \ref{table:textvqa}, \ref{table:stvqa} show that our large size model outperforms the comparably sized previous state-of-the-art method LaTr \cite{biten2022latr} by \textcolor{forestgreen}{+3.4\%}, \textcolor{forestgreen}{+2.4\%} and \textcolor{forestgreen}{+2.2\%} on the OCR-VQA, TextVQA, and ST-VQA test sets respectively.
These results show that our method  generalizes beyond document understanding tasks.

\textbf{Analysis:} Surprisingly, on OCR-VQA,  \papertitlenospace$_{\text{large}}$ even performs better than GIT2 \cite{wang2022git} which is a 5.1B size model (\vs 750M for \papertitlenospace$_{\text{large}}$) and uses 12.9B data for pre-training (\vs 64M for \papertitlenospace$_{\text{large}}$).
On TextVQA, \papertitlenospace \ does better than several vision-language models which are much bigger and have been pre-trained on much more data. On the test set, it is (\textcolor{forestgreen}{+9.9\%}) better than Flamingo (which at 80B has 106x the number of parameters). On the validation set, it is better than PaLi-3B and 15B (\textcolor{forestgreen}{+2.2\%}, \textcolor{forestgreen}{+6.8\%}) respectively. GIT2 and PaLi-17B do perform better than it. (GIT2 also uses 8 VQA datasets to train). \papertitle gets this performance \uline{without} any natural image-text pre-training. We present this as evidence that  \papertitle is a good approach to solving this problem with a much smaller model and much less data.

\section{Ablation Experiments}
\label{sec:experiments:ablation_experiments}

\noindent \textbf{\textit{Ablation of \papertitleshort novel pre-training tasks}}: Table \ref{table:ablation:pretrain_tasks} shows \papertitleshort ablation on the proposed novel pre-training tasks and multi-modal training. The denoising language modeling task and spatial features mentioned in \S \ref{sec:approach:arch:spatial} are  applied to all architectures. Note, this ablation was performed on \papertitleshort-small with 1M doc  pre-training. 

\vspace{-2mm}
\begin{table}[htbp]
\centering
\scalebox{0.75}{ 
\begin{tabular}{l|HcHH} 
\multirow{2}{*}{Model Ablation}  & \multicolumn{4}{c}{Datasets}  \\  
& \thead{FUNSD\\(F1)} & \thead{DocVQA (ANLS)} & \thead{STVQA\\(eval Acc.)} & \thead{TextVQA\\(eval Acc.)}   \\ 
\Xhline{2.0\arrayrulewidth} 
baseline B & 82.7 & 69 & - & -\\   
B + V & 83.3  & 70.5 \textcolor{forestgreen}{(+1.5)} & 64.9 & 55.8 \\   
B + V + L & 82.5 & 71.2 \textcolor{forestgreen}{(+2.2)} & 64.9 & 55.4 \\   
B + V + G & 82.7 & 71.7 \textcolor{forestgreen}{(+2.7)} & 64.4 (ongoing) & 54.3 (ongoing) \\  
B + V + L + G & 83.0 & 73.0 \textcolor{forestgreen}{(+4.0)} & 64.2 (ongoing) & 53.9 (ongoing) \\   
\Xhline{2.0\arrayrulewidth} 
\end{tabular} 
}
\caption{\textbf{\papertitle Pre-training Tasks Ablation}: Impact of three pre-training tasks on four downstream tasks over baseline. \textbf{B:} baseline, \textbf{V:} only with Visual features \S \ref{sec:approach:arch:visual}, \textbf{L:} with Token-to-Line prediction pre-training \S \ref{sec:approach:pretrain:tok_to_line}, \textbf{G:} with Token-to-Grid prediction pre-training \S \ref{sec:approach:pretrain:tok_to_grid}.  
}
\label{table:ablation:pretrain_tasks} 
\end{table}

\begin{table}
\centering 
\scalebox{0.75}{ 
\begin{tabular}{l|l|ll} 
\multirow{3}{*}{Model} & { \# pre-train data} & \multicolumn{2}{c}{Datasets}  \\  
& & \thead{FUNSD } & \thead{CORD }    \\ 
\midrule
LayoutLMv2$_{\text{base}}$ \cite{xu2020layoutlmv2} & 11M & 82.7 & 94.9 \\
\papertitlenospace$_{\text{base}}$ & 11M & 86.1 \textcolor{forestgreen}{(+3.4\%)} & 96.2 \textcolor{forestgreen}{(+1.3\%)} \\   
\papertitlenospace$_{\text{base}}$ & 64M & 87.9 \textcolor{forestgreen}{(+5.2\%)} & 96.8 \textcolor{forestgreen}{(+1.9\%)}\\  
\papertitlenospace$_{\text{base}}^{\dagger}$&64M&88.3 \textcolor{forestgreen}{(+5.6\%)}&96.8 \textcolor{forestgreen}{(+1.9\%)}\\  
\end{tabular} 
}
\caption{\textbf{\papertitle Pre-training Data Ablation}: Impact of training with different \# of pre-training data on various down-stream tasks. The F1 scores are reported. $^{\dagger}$ indicates the combination of the ST-VQA and TextVQA training sets is used.}
\label{table:ablation:pretrain_data} 
\vspace{-3mm}
\end{table}

 \vspace{-3mm}
\noindent \textbf{Pre-training Impact or Better Approach?} \papertitleshort was pre-trained with 64M documents whereas prior-art 
like LayoutLMv2 \cite{xu2020layoutlmv2} 
was pre-trained with only 11M documents.
In order to see if \papertitleshort benefits come from more pre-training data or a better approach, we ablate. Table \ref{table:ablation:pretrain_data} shows that \papertitleshort$_{\text{base}}$ is superior to LayoutLMv2$_{\text{base}}$ when pre-trained on the same quantity of data. We also see \papertitleshort improve in performance as more pre-training data is provided (64M). The table shows that the novel \papertitleshort asymmetric pre-training approach is a superior VDU approach.

\begin{figure}[!h]
  \centering
  \includegraphics[width=0.9\linewidth]{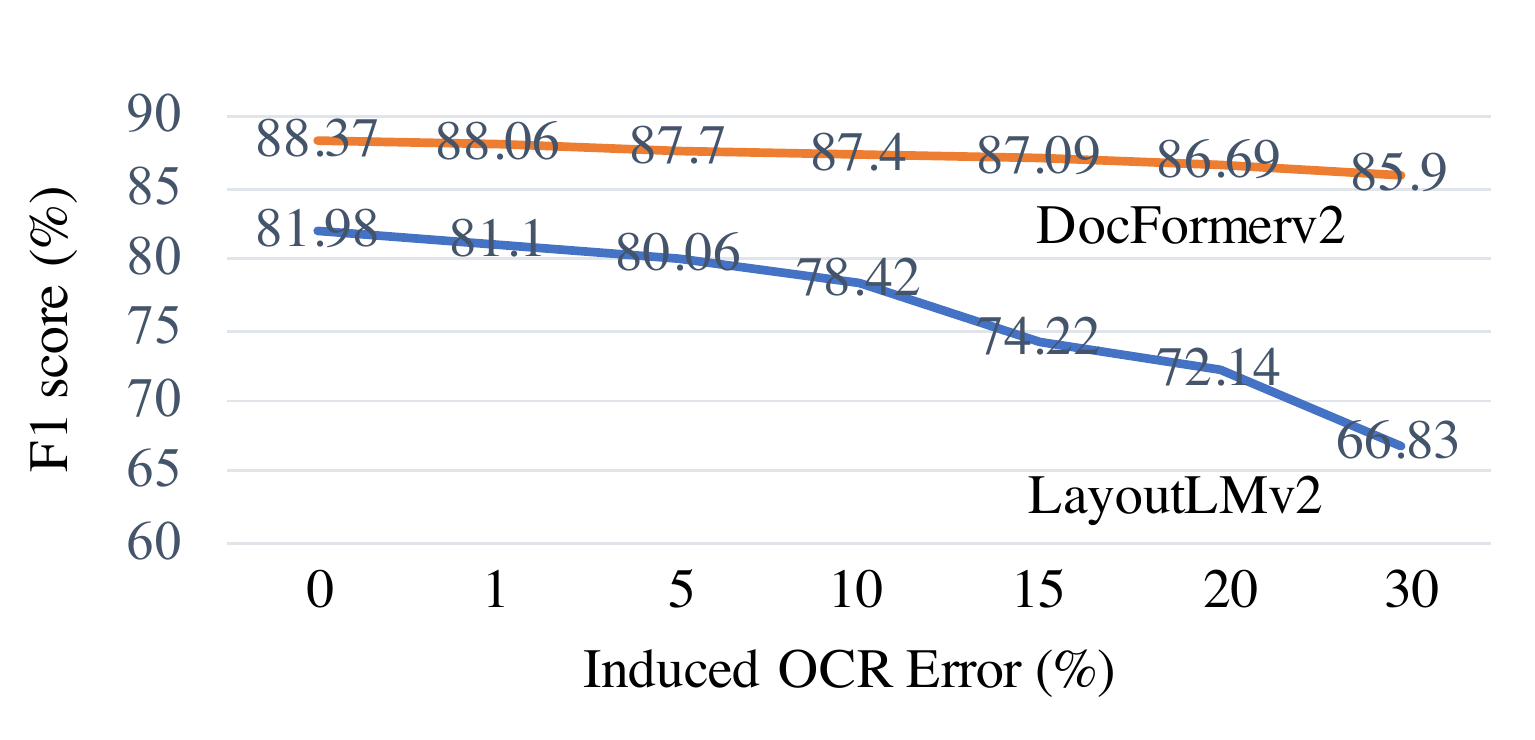}
  \caption{\textbf{Induced OCR Error Ablation}. F1 score performance evaluated on FUNSD for varying orders of injected OCR errors.
  }
  \label{fig:ocrerror}
  \vspace{-3mm}
\end{figure}

\noindent \textbf{\textit{Robustness to OCR errors}.} \papertitleshort consumes OCR text which can have errors. Since it also has a generative decoder, in theory it is robust to certain distortions and noise from OCR-text. To quantify the degree of robustness, we conduct a study using the FUNSD dataset and artificially introduce noise/typographical errors to the input words, simulating OCR errors. Specifically, for every character in the text, we randomly replace it with an erroneous character with a probability $p$, limiting to a maximum of 1 character error per word. We then evaluate the performance of the DocFormerv2\textsubscript{base} and LayoutLMv2\textsubscript{base} models on the error injected text to observe their resilience to such noise. From Figure \ref{fig:ocrerror}, we see that for increasing amount of injected OCR errors. Specifically, 20\% OCR errors only decreases the performance by -1.68\% whereas an encoder-only model decreases by a wide margin -9.84\%. This shows the benefit of our approach (in having a generative decoder).


\vspace{-5mm}



\begin{figure}[t]
  \centering
  \begin{minipage}[b]{0.4 \linewidth}
         \centering
         \includegraphics[width=1.2\linewidth]{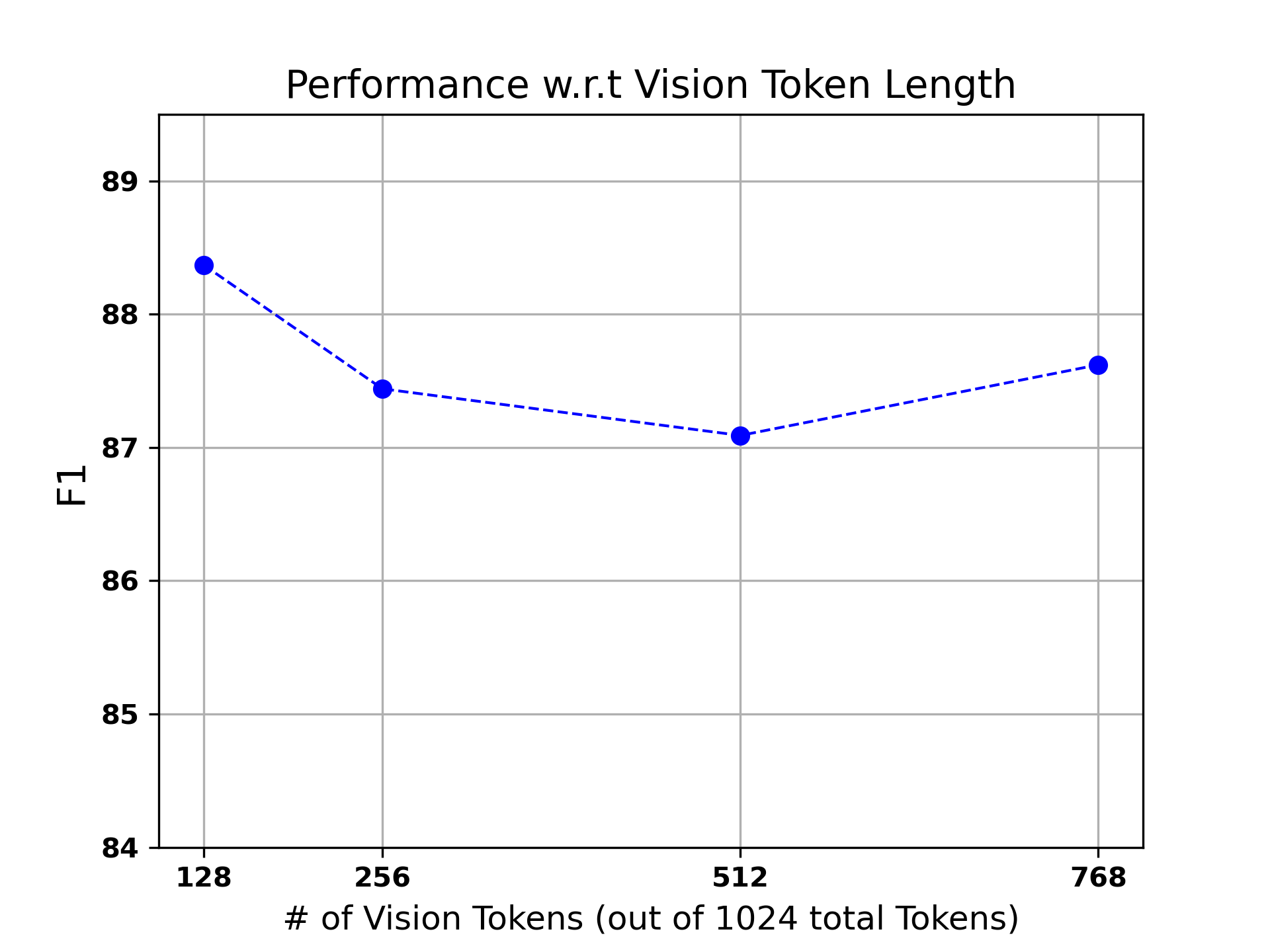}
         \caption{\textbf{Image token length ablation}. Effect on model performance w.r.t variation of the proportion of vision tokens to text tokens provided as input to the model. 128 works best and was used as the final model design.}
         \label{fig:ablation:vision_seq_len}
     \end{minipage}
     \hspace{0.02\linewidth}
     \begin{minipage}[b]{0.4 \linewidth}
         \centering
         \includegraphics[width=1.2\linewidth]{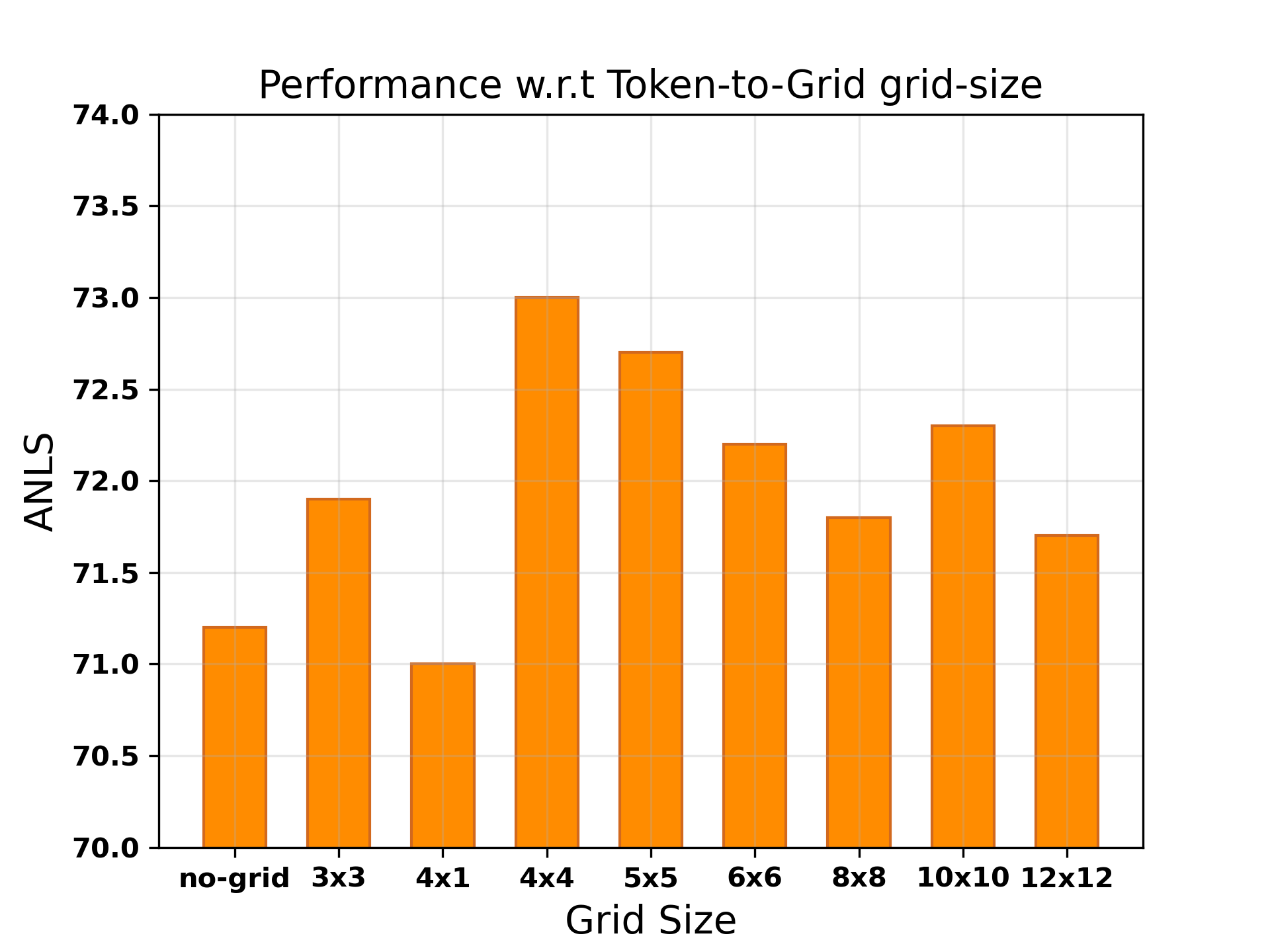}
         \caption{\textbf{Token-to-Grid Ablation}. How different grid sizes used for the Token-to-Grid pre-training task affects model performance on DocVQA. 4x4 seems best and was used for all final pre-training.}
         \label{fig:ablation:token_to_grid}
     \end{minipage}

\vspace{-5mm}
\end{figure}

\noindent \textbf{\textit{Varying Image Tokens.}} Concatenating the image tokens along with the text tokens is a simple and intuitive approach for the model to learn to jointly capture multi-modal information. However, since we are limited on the total number of tokens we can use, it begs the question - what is a suitable proportion of vision-to-text tokens to be used for this design? We perform ablations in this regard to measure the performance obtained by finetuning the model on FUNSD dataset with varying ratios of vision tokens to text tokens. Figure \ref{fig:ablation:vision_seq_len} shows that $128$ image tokens appears to provide the best performance compared to the other settings. 



\noindent \textbf{\textit{Do we need a separate Image encoder?}} We investigate if \papertitle's way of consuming visual features is optimal for VDU. Instead of using linear features, we use Swinv2 \cite{Liu2021Swinv2} and pre-train this setup on 1M documents. When fine-tuned on DocVQA we observe that the setup with Swinv2 as visual backbone, substantially under-performs our approach by (+4.3\%). For this task, the complex visual features from Swinv2 are less beneficial than our simple linear features.


\begin{table}[htbp]
\centering
\scalebox{0.81}{
\begin{tabular}{l|l|l}
  Model & image encoder & \thead{DocVQA \cite{mathew2020docvqa} eval ANLS (\%)} \\
  \midrule
  baseline & - & 69.0 \\
  \papertitlenospace$_{\text{small}}$ & Swinv2${_{\text{small}}}$ \cite{Liu2021Swinv2} & 66.2 \\
  \papertitlenospace$_{\text{small}}$ & Linear (ours) & 70.5 \textcolor{forestgreen}{(+4.3\%)}  \\
\end{tabular}
}
\caption{\textbf{Image Encoder Ablation}: All models pre-trained on 1M docs from IDL. Swinv2 too was pre-trained and fine-tuned.}
\label{table:ablation:image_encoder}
\vspace{-0.3cm}
\end{table}

\noindent\textbf{Correct grid size for Token-to-Grid pre-training?} In \S \ref{sec:approach:pretrain:tok_to_grid}, we presented the novel Token-to-Grid pre-training task. In this pre-training ablation \S \ref{table:ablation:pretrain_tasks} this task was observed to provide benefits. Here the appropriate virtual grid-size is empirically determined. From Fig. \ref{fig:ablation:token_to_grid}, 4x4 grid seems optimal. Smaller or asymmetric grid structures (4x1) seem to cause harm. On the other end, if the grid is too granular (12x12, 8x8), the performance seems to hurt as well. All models pre-trained on \papertitleshort$_{\text{small}}$ and 1M documents from IDL, with the Vision and Token-to-line enabled. 

\section{Conclusion}

Our work \papertitle highlights the importance of two novel pre-training tasks and the efficacy of enriching encoder representations with local semantic information via pre-training tasks. We perform experiments on eight varied datasets (five on VDU and three on scene-text VQA) achieving state-of-the-art numbers on all datasets. Based on ablations, we also show the various 
design choices and its impact on downstream performance.







{\small
\bibliographystyle{ieee_fullname}
\bibliography{egbib}
}

\end{document}


\title{DocFormerv2 Supplemental Paper}

\author{Srikar Appalaraju$^{1}$ \thanks{Corresponding author.}, \quad Peng Tang$^{1}$ , \quad Qi Dong$^{1}$, \quad Nishant Sankaran$^{1}$, Yichu Zhou$^{2}$ \thanks{The research was conducted during Y. Zhou  internship at AWS.}, \quad R. Manmatha$^{1}$ \\
$^{1}$ AWS AI Labs \quad
$^{2}$ School of Computing at University of Utah\\
{\tt\small \{srikara, tangpen, qdon, nishsank, manmatha\}@amazon.com, flyaway@cs.utah.edu}
}

\maketitle
\ificcvfinal\thispagestyle{empty}\fi

\newcommand{\xrightarrowdbl}[2][]{%
	\leftarrow\mathrel{\mkern-14mu}\xrightarrow[#1]{#2}
}
\newcommand{\papertitle}{DocFormerv2 }
\renewcommand\theadset{\def\arraystretch{.85}}

\section{Supplemental}


This is the supplemental material for the main \papertitle paper 
Please read the main paper for model formulation, performance numbers on various datasets and further analysis and ablation. \papertitle achieves state-of-the-art performance on eight datasets each with strong baselines. Specifically, on Table Fact \cite{Chen2019TabFactAL, Borchmann2021DUE} \textcolor{forestgreen}{(+4.0\%)}, InfoVQA \cite{mathew2022infographicvqa} (1.4\%), FUNSD \cite{Jaume2019FUNSDAD} (1\%). On three VQA tasks involving scene-text, \papertitle outperforms previous comparably-sized models and even does better than much larger models (such as GIT2, PaLi and Flamingo) on some tasks.

\section{Table VQA - WikiTableQuestions dataset}
We further show performance on \papertitle on WikiTableQuestions \cite{Pasupat2015CompositionalWTQMetric, Borchmann2021DUE} dataset. This a challenging dataset requiring complex spatial reasoning and arithmetic (counting). See \S \ref{sec_supple:wikivqa} for visualizations. As can be seen in Table \ref{table_supp:tablevqa}, \papertitle achieves state of the art performance of 48.2 Accuracy on this challenging dataset.

\begin{table}[htbp]
\centering
\scalebox{0.81}{
\begin{tabular}{l|c|c|cH}
  Model & \thead{pre-train\\data} & \thead{\#param} & \thead{\textbf{WikiTableQuestions} \\Acc. (\%)} \\
  \midrule
  \multicolumn{5}{l}{\textit{methods based on only text / (text + spatial) features:}} \\ 
  \midrule
  T5$_{\text{large}}$\cite{Raffel2019ExploringT5} & - & 750M & 33.3 \\
  T5$_{\text{large}}$+U \cite{Borchmann2021DUE} & - & 750M & 38.1\\
  T5$_{\text{large}}$+2D \cite{Borchmann2021DUE} & - & 750M & 30.8 \\
  T5$_{\text{large}}$+2D+U \cite{Borchmann2021DUE} & - & 750M & 43.3 \\
  \midrule
  \multicolumn{5}{l}{\textit{methods based on image + text + spatial features:}} \\ 
  \midrule
  LayoutLMv3$_{\text{large}}$ \cite{huang2022layoutlmv3} & 11M & 368M & 45.7 \\
  UDOP \cite{Tang2022UnifyingUDOP} & 11M & 794M & \underline{47.2} \\
  \midrule
  DocFormerv2$_{\text{large}}$ & 64M & 750M & \textbf{48.3} \textcolor{forestgreen}{(+1.1\%)}  \\
\end{tabular}
}
\caption{\textbf{Comparison on Table VQA Dataset - WikiTableQuestions }\cite{Borchmann2021DUE}: Our work, \papertitle outperforms the previous state of the art. 
}
\label{table_supp:tablevqa}
\vspace{-0.4cm}
\end{table}

\section{Implementation Details} We present all the hyper-parameters in Table \ref{table:implementation_full_1} used for pre-training and fine-tuning \papertitle. We follow the standard practice of unsupervised pre-training followed by supervised fine-tuning \cite{xu2020layoutlm,xu2020layoutlmv2,bros2020hong,appalaraju2021docformer,biten2022latr,Ho2022YOROL}. We emphasize that even for Text-VQA datasets, the \uline{same} document pre-trained model was used. We only performed supervised fine-tuning on provided Text-VQA train data.

Specifically, FUNSD \cite{jaume2019} datasets were fine-tuned for 300 epochs. CORD \cite{park2019cord} for 200 epochs. 

\begin{table}[H]
	\begin{center}
	\scalebox{0.75}{
		\begin{tabular}{l|c|c}
			Hyper-Parameter & Pre-training & Fine-tuning \\
			\Xhline{2.5\arrayrulewidth}
			Epochs & 3 & varies \\
			Learning rate & 5E-05 & 1E-05/2.5E-05 \\
			Warm-up & 1000 iters & 0 \\
			Gradient Clipping & 1.0 & 1.0 \\
			Gradient agg. & False & False \\
			Optimizer & AdamW\cite{Loshchilov2017DecoupledWD} & AdamW\cite{Loshchilov2017DecoupledWD} \\
			Lower case & True & True  \\
			Sequence length & 512 & varies \\
			Encoder layers & 12/24 (B/L) & 12/24 (B/L) \\
			32-bit mixed precision & True & True \\
			Batch size & 9/1 per GPU (B/L) & 9/1 per GPU (B/L) \\
			GPU hardware & A100 (40GB) & A100 (40GB) \\
			Training Num. Samples & 64M & varies \\
		\end{tabular}
		}
	\end{center}
	\vspace{-3ex}
	\caption{\textbf{Implementation Details}: Hyper-parameters used for pre-training \papertitle and fine-tuning for downstream tasks. Training epochs vary for down-stream tasks. \textbf{B}: \papertitle$_{\text{base}}$ and \textbf{L}: \papertitle$_{\text{large}}$  }
	\label{table:implementation_full_1}
	\vspace{-1ex}
\end{table}

For pre-training, the \papertitle is initialized with T5 \cite{raffel2019exploring} pre-trained weights for the language branch and keep the visual weights randomly initialized. During unsupervised pre-training, the \papertitle learns to fuse and utilize multi-modal features (vision, language and spatial) to minimize the three objectives - Token-to-Grid, Token-to-Line and denoising language modeling task.

\textbf{DUE Benchmarks}\cite{Borchmann2021DUE}: Is a collection of seven datasets on document understanding task. We use four datasets from the DUE benchmarks - DocVQA \cite{mathew2020docvqa}, InfoVQA \cite{mathew2022infographicvqa}, TabFact \cite{Chen2019TabFactAL} and WikiTableQuestions \cite{Pasupat2015CompositionalWTQMetric}. They were chosen due to their diversity and to show the versitility of \papertitle. The other three datasets DeepForm , KleisterCharity and PWC are similar to VQA setting and in the interest of time, we do not evaluate on them. 

\section{Datasets}
In this section we describe the pre-training and fine-tuned datasets used in \papertitle paper.

\subsection{Pre-training Dataset}
Following prior-art \cite{biten2022latr, appalaraju2021docformer, Tang2022UnifyingUDOP, xu2020layoutlm, xu2020layoutlmv2, huang2022layoutlmv3} we pre-train \papertitle on Industrial Document Library (IDL) dataset \footnote{https://www.industrydocuments.ucsf.edu/}. As documented in the main paper, the IDL is an electronic repository of documents generated by industries that have an impact on public health, and is hosted by the University of California, San Francisco Library. It encompasses millions of publicly disclosed documents from a range of industries such as pharmaceuticals, chemicals, food, and fossil fuels. The IDL dataset, which contains approximately 13 million documents, equating to about 70 million pages (64 million usable) of various document images, was crawled from the website. Various document types, such as forms, tables, and letters, with diverse layouts are available on IDL. To extract text from the documents, OCR model was run for each document. OCR predictions could be noisy (depending on the quality of the document image). Before pre-training, the crawled and OCR’ed IDL data was pre-processed. Pre-processing was done to discard documents based on quality. After pre-processing and pruning, 64 million documents (approximately 6 million documents were discarded) were retained for pre-training. 

\begin{figure}[htbp]
	\centering
    {\includegraphics[width=0.8\linewidth]{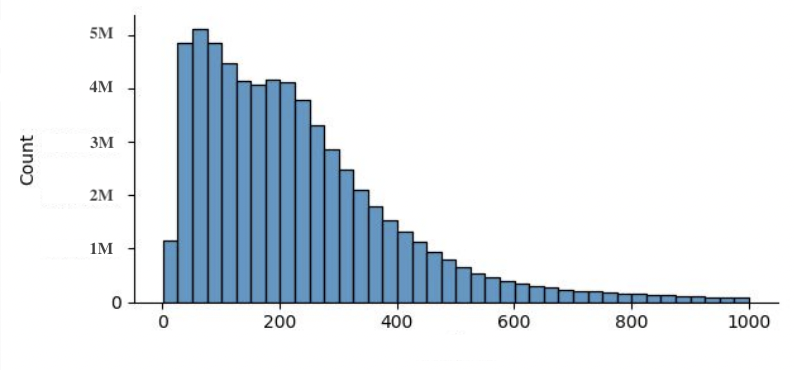}}
	
	\caption{\textbf{IDL data distribution}: Frequency plot for predicted OCR tokens in IDL dataset used for \papertitle pre-training} 
	\label{fig:plot_idl}
\end{figure}

\begin{figure}[htbp]
\centering
\begin{minipage}{0.3\textwidth}
\centering
\includegraphics[width=1\linewidth]{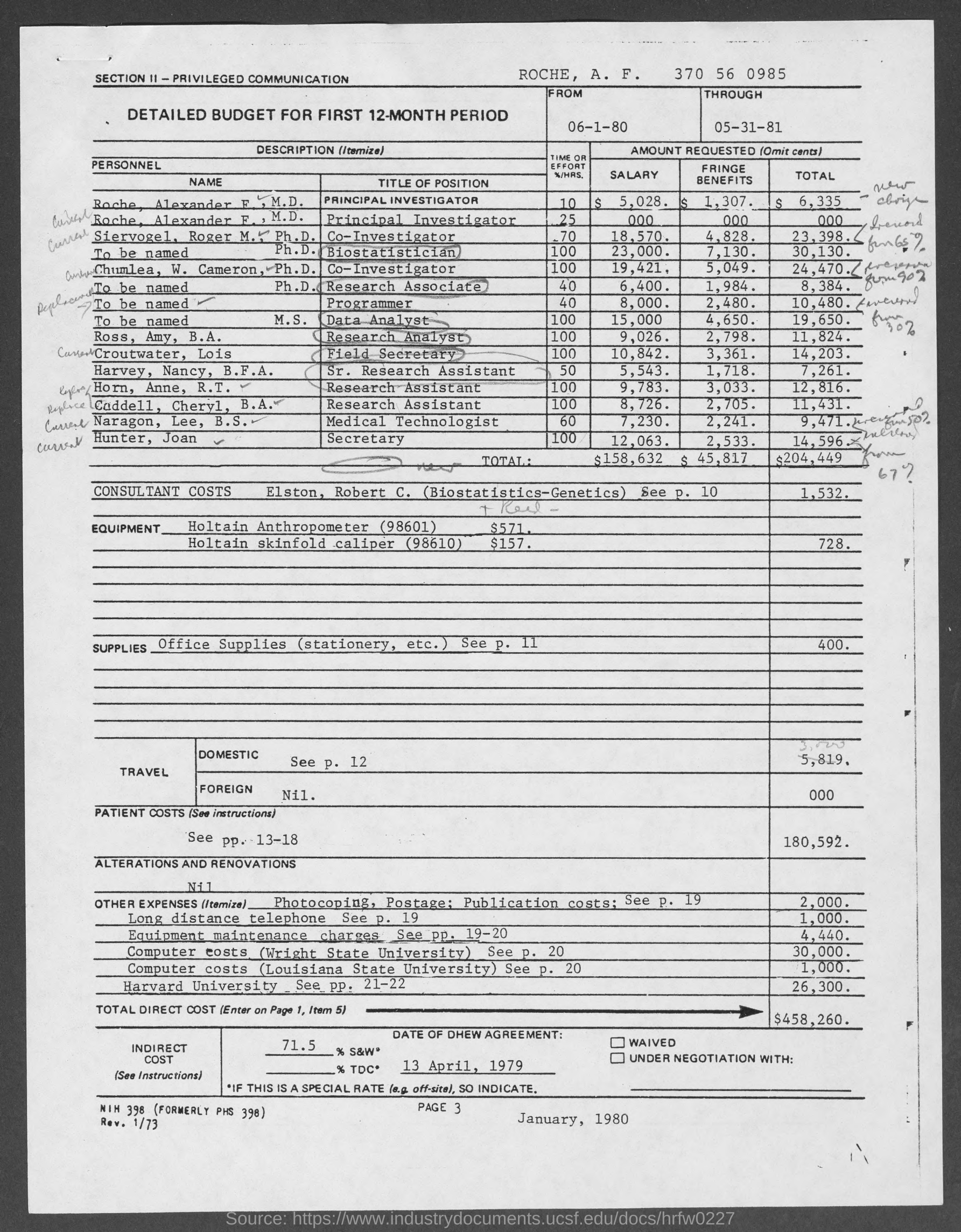}
\label{fig:image1}
\end{minipage}
\hfill
\begin{minipage}{0.3\textwidth}
\centering
\includegraphics[width=1\linewidth]{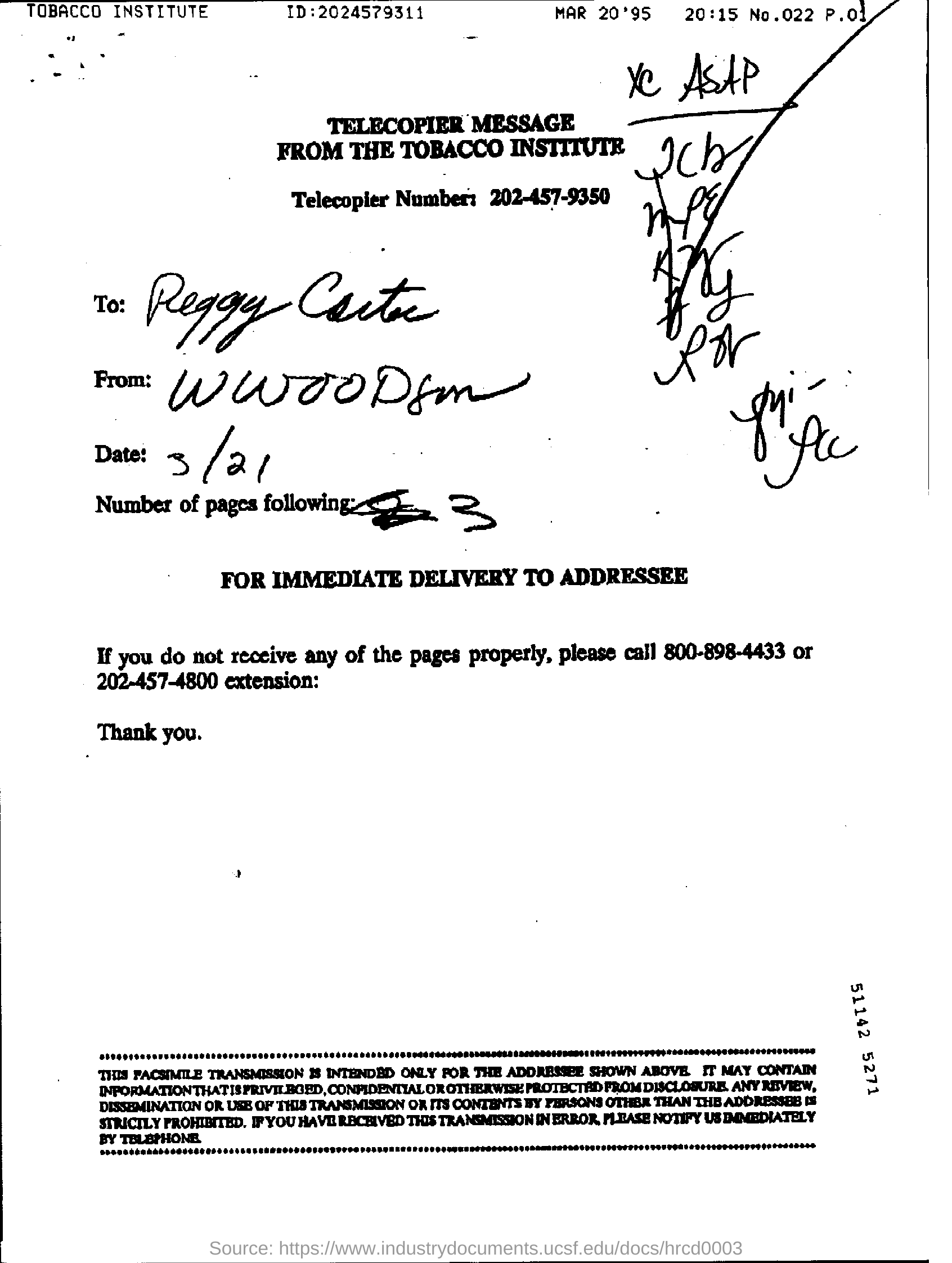}
\label{fig:image2}
\end{minipage}
\hfill
\begin{minipage}{0.3\textwidth}
\centering
\includegraphics[width=1\linewidth]{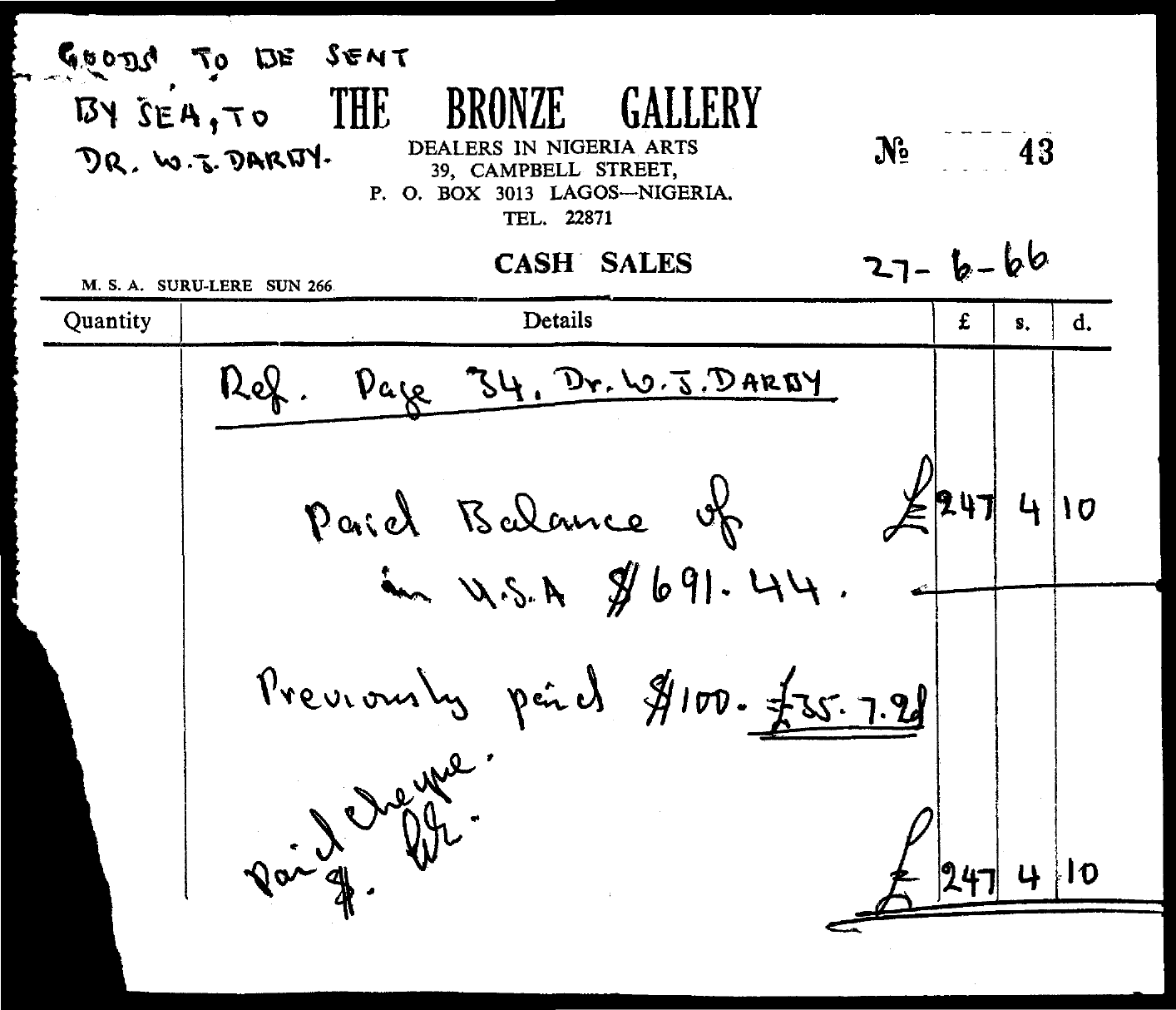}
\label{fig:image3}
\end{minipage}
\caption{\textbf{IDL sample documents visualization}} 
\label{fig:idl_viz}
\end{figure}

The detected OCR word distribution across all the 64M documents is presented in Fig. \ref{fig:plot_idl}, indicating a right-skewed normal distribution with the majority of documents containing 20 to 400 words per document. We note that IDL dataset contains more words on average than image-text datasets like CC  and LAION datasets, which is particularly advantageous for pre-training in Text-VQA tasks. Fig. \ref{fig:idl_viz} illustrates representative examples from the IDL dataset.

\subsection{Document Understanding Datasets}

\begin{table}[h]
\begin{center}
\scalebox{0.75}{
\begin{tabular}{l|lll|l|l}
\multicolumn{1}{c|}{\multirow{2}{*}{Dataset}} & \multicolumn{3}{l|}{Size (k documents)} & \multirow{2}{*}{Type} & \multirow{2}{*}{Metric} \\
\multicolumn{1}{c|}{}                         & Train        & Dev         & Test       &                       &                         \\ \Xhline{2.5\arrayrulewidth}
TabFact \cite{Chen2019TabFactAL}               & 13.2         & 1.7         & 1.7        & Table NLI             & Acc.                    \\
WikiTableQuestions \cite{Pasupat2015CompositionalWTQMetric}       & 1.4         & 0.3         & 0.4        & Table QA             & Acc.                    \\
DocVQA  \cite{mathew2020docvqa}          & 10.2         & 1.3         & 1.3        & Visual QA             & ANLS                    \\
InfoVQA  \cite{mathew2022infographicvqa}       & 4.4          & 0.5         & 0.6        & Visual QA             & ANLS                    \\
FUNSD  \cite{Jaume2019FUNSDAD}                 & 0.15        & -           & 0.05       & Entity Labeling       & F1                      \\
CORD   \cite{park2019cord}                 & 0.8          & 0.1         & 0.1        & Entity Extraction     & F1                      \\
\end{tabular}
}
\caption{\textbf{Document Understanding Datasets Details}: Dataset types and their train/val/test split distributions. }
\label{table:doc_und_dataset_details}
\end{center}
\end{table}

\textbf{TabFact} \cite{Chen2019TabFactAL} contains 16k Wikipedia tables as the basis for 118k human annotated questions and statements sourced through Amazon Mechanical Turk (AMT). The dataset focuses on the task of Natural Language Inference (NLI) based on reasoning with the provided tabular data. The goal is to recognize which statements are validated/refuted by the tabular content associated with it.

\textbf{WikiTableQuestions} \cite{Borchmann2021DUE, Pasupat2015CompositionalWTQMetric} is a dataset consisting of 2,108 semi-structured HTML tables from Wikipedia and 22,033 question-answer pairs about the content in the tables. AMT workers gathered trivia questions about the tables that covered a wide range of operations such as table lookup, aggregation, joins, etc.

\textbf{DocVQA} \cite{mathew2020docvqa} consists of 50,000 questions defined on 12,767 document
images. The task it defines is to verify the capability of answering questions about document images and the textual content within. It encompasses typed and handwritten document samples and also requires the comprehension of layout, indicators (like ticks, etc) and style of the presented content to answer questions about it.

\textbf{InfographicVQA} \cite{mathew2022infographicvqa} is a collection of 5,485 images about infographics collected from 2,594 web domains. It poses 30,035 questions pertaining to these images grounded on tables, figures and visualizations. The questions involve basic numeric reasoning elements such as counting, arithmetic and sorting operations.

\textbf{FUNSD} \cite{Jaume2019FUNSDAD} is a dataset that poses the challenge of understanding forms structure and content from noisy scanned images. It consists of 199 real, fully annotated scanned forms which are split into 149 for train and 50 for test. It has defines 3 tasks: (i) word grouping (clubbing words belonging to the same semantic entity), (ii) semantic entity labeling and (iii) entity linking (establish relations between entities).

\textbf{CORD} \cite{park2019cord} consists of 11,000 scanned receipts collected from shops and restaurants which contains annotations for OCR and multi-level semantic parsing tasks. The fine-grained classification task entails 8 super-classes and 54 subclass labels for the word entities within the receipts like menu name, quantity, total price, and so on.

\subsection{Text-VQA datasets}

\textbf{TextVQA}\cite{singh2019towards} contains 28k images from the Open Images dataset. The questions and answers are collected
through Amazon Mechanical Turk (AMT). The workers are instructed to come up with questions that require
reasoning about the scene text in the image using the following process - , 10 answers were collected for each question.

\textbf{ST-VQA }\cite{biten2019scene} is an aggregation of well-known computer vision datasets, namely: ICDAR 2013 , ICDAR2015 , ImageNet , VizWiz [7], IIIT Scene Text Retrieval , Visual Genome  and COCO-Text .
ST-VQA is collected through Amazon Mechanical Turk, asking workers to
come up with questions so that the answer is always the
scene text in the image. 

\textbf{OCR-VQA} \cite{mishra2019ocr} is composed of 207,572 images of book
covers and contains more than 1 million QA pairs about these images. 
The questions are template-based, 
asking about information on the book such as title, author,
year. The questions are all can be answered by inferring the
book cover images.

\begin{table}[t]
\centering
\resizebox{\linewidth}{!}{
\begin{tabular}{l|c|c|c}
  Dataset & Train Set & Val Set & Test Set\\
  \midrule
  OCR-VQA \cite{mishra2019ocr} & 166K/801.7K & 20.7K/100K & 20.8K/100.4K \\
  TextVQA \cite{singh2019towards} & 21.9K/34.6K & 3.2K/5K & 3.3K/5.7K \\
  ST-VQA \cite{biten2019scene} & 17K/23.4K & 1.9K/2.6K & 3K/4.1K \\
\end{tabular}
}
\caption{\textbf{Text-VQA Dataset Stats:} The number of images/questions of different text-VQA datasets. We report performance on Test-set in the main paper after training only on Train-set.}
\label{table_supp::datasets}
\end{table}

\begin{table*}[H]
	\begin{center}
\begin{tabular}{cccccccccc}
\multirow{2}{*}{ Task } & \multicolumn{3}{|c|}{ Size (k documents) } & \multirow{2}{*}{$\begin{array}{l}\text { Mean samples } \\
\text { per document }\end{array}$} & \multirow{2}{*}{ Type } & \multirow{2}{*}{ Metric } & \multicolumn{2}{|c|}{ Features } & \multirow{2}{*}{ Domain } \\

\hline & Train & Dev & Test & & & & Input & Scanned & \\
\hline DocVQA & 10.2 & 1.3 & 1.3 & 3.9 & Visual QA & ANLS & \multirow{5}{*}{ Doc. } & + & Business \\
\hline InfographicsVQA & 4.4 & 0.5 & 0.6 & 5.5 & Visual QA & ANLS & & - & Open \\
\hline Kleister Charity & 1.7 & 0.4 & 0.6 & 7.8 & KIE & F1 & & $+1-$ & Legal \\
\hline PWC & 0.2 & 0.06 & 0.12 & 25.5 & KIE & ANLS $^2$ & & - & Scientific \\
\hline DeepForm $\star$ & 0.7 & 0.1 & 0.3 & 4.8 & KIE & $\mathrm{F} 1$ & & $+1-$ & Finances \\
\hline WikiTableQuestions ${ }^{\star}$ & 1.4 & 0.3 & 0.4 & 11.3 & Table QA & Acc. & \multirow[b]{2}{*}{ Exc. } & - & Open \\
\hline $\mathrm{TabFact}^{\star}$ & 13.2 & 1.7 & 1.7 & 7.1 & Table NLI & Acc. & & - & Open \\
\hline
\end{tabular}
\end{center}
\end{table*}
\begin{figure*}[th]
	\centering
    {\includegraphics[width=0.9\linewidth]{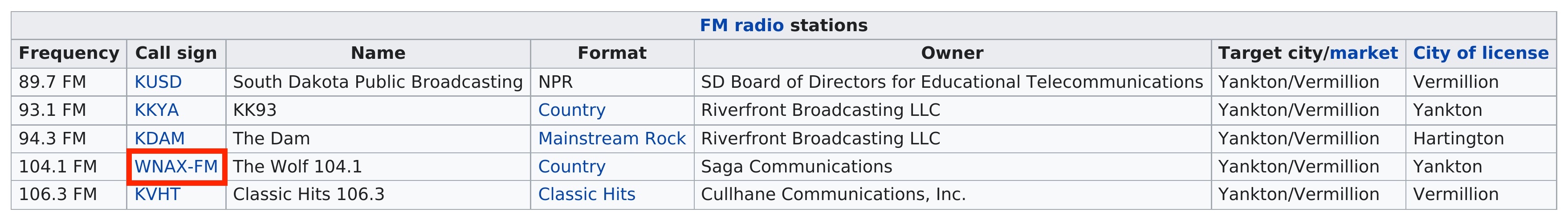}}
	\caption{\textbf{\papertitle predictions on WikiTableQuestions}: For the question \"which of these stations do not have a 'k' in their call sign?\", \papertitle correctly predicts \"WNAX-FM\" (4th row, 2nd column). This particular example requires reasoning over spatial, visual and language features. Image file in test: csv\_200\-csv\_18\_0\.jpeg. The \textcolor{red}{\textbf{Red}} box showing the answer for locating the answer.}
	\label{fig:wtq_supplemental_viz1}
\end{figure*}
\begin{figure*}[!h]
	\centering
    {\includegraphics[width=0.9\linewidth]{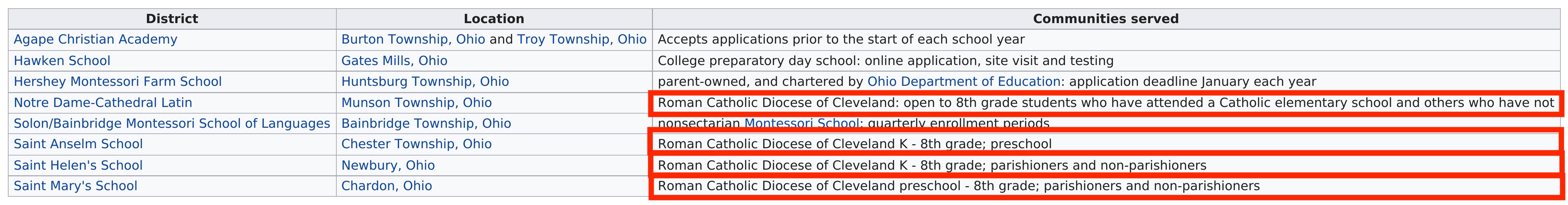}}
	\caption{\textbf{\papertitle predictions on WikiTableQuestions}: For the question \"how many of the schools serve the roman catholic diocese of cleveland?\", \papertitle correctly predicts \"4\". This particular example requires arithmetic counting and reasoning over multiple rows as the model needs to count to generate the final answer. We point to the reader that no arithmetic specific pre-training like Tapas \cite{herzig2020tapas} was done for \papertitle. Image file in test: csv\_200\-csv\_8\_0\.jpeg}
	\label{fig:wtq_supplemental_viz2}
\end{figure*}
\begin{figure*}[!h]
	\centering
    {\includegraphics[width=0.9\linewidth]{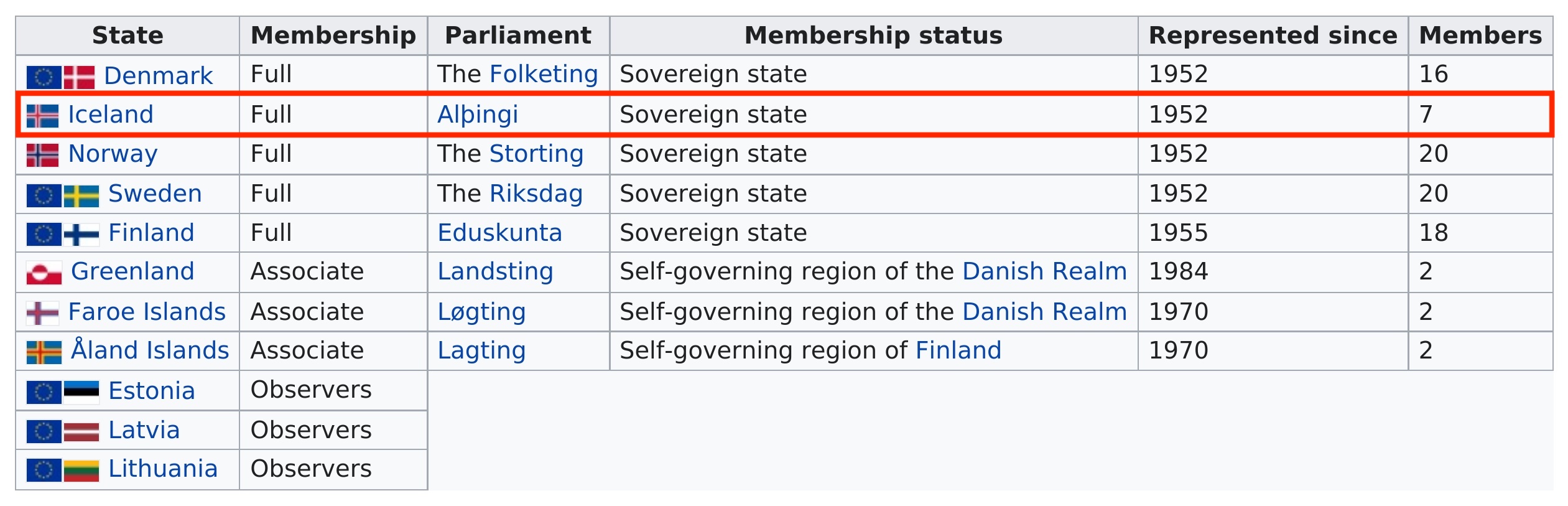}}
	\caption{\textbf{\papertitle predictions on WikiTableQuestions}: For the question \"which state has a full membership and also has a membership status under sovereign state with only 7 members?\", \papertitle correctly predicts \"Iceland\". We hypothesize that our Token-to-Line pre-training helps in such scenarios as our approach has efficiently learned to make sense of local information (here information in a row). Image file in test: csv\_201\-csv\_20\_0.jpeg}
	\label{fig:wtq_supplemental_viz3}
\end{figure*}
\section{Related Work Discussion}
We add more discussions of the existing works in Visual Document Understanding domain.

\noindent\textbf{Encoder-Only VDU models} LayoutLMv1\cite{xu2020layoutlm} is a representative work using a Transformer encoder model to tackle the entity extraction and labelling task. Besides text level token embeddings, LayoutLMv1 adds spatial embeddings and OCR-patch visual embedding for individual word tokens. Specifically, it uses OCR box locations as 2D position embeddings, and uses a Faster-RCNN model to extract local visual features for each input word. The final embedding of each input tokens are the summation of text embedding, 2D position embedding, 1D position embedding and OCR patch visual embedding. To enhance the visual information for each token, LayoutLMv2\cite{xu2020layoutlmv2} uses one CNN to extract the global image features, and splits the image features into multiple visual tokens, and concatenates them with text tokens as the input sequence. It emphasises the visual information learning in attention layers. To further simplify the visual token representation, LayoutLMv3 \cite{huang2022layoutlmv3} uses one Conv2D layer to split the original images into multiple visual tokens, rather than using an extra visual encoder to extract global image features, and leveraging visual features as visual tokens. These encoder-only models provide simple frameworks to handle VDU tasks, but limit the application scenarios due to the token classification formulation. For example, it is hard to obtain the complete entities by leveraging the predicted BIO tags for non-reading order input sequences and complex entity layouts. 

\noindent\textbf{Encoder-Decoder VDU models} TILT \cite{powalski2021going} designs a Transformer Encoder-Decoder model to tackle entity labelling and entity extraction tasks. It uses U-Net as visual encoder and uses T5 decoder to conduct the label prediction. While TILT \cite{powalski2021going}  proposed a encoder-decoder transformer for VDU, they only train it on one pre-training task (masked language modeling) and also use a bulky visual CNN. Our approach \papertitle , not only simplifies the architecture by not using a separate visual module (CNN or Transformer based) and has multiple unsupervised pre- training tasks.

\section{Qualitative Analysis}
In this section, we present qualitative analysis of \papertitle by visualizing its predictions on various datasets.

\begin{figure*}
	\centering
    {\includegraphics[width=0.9\linewidth]{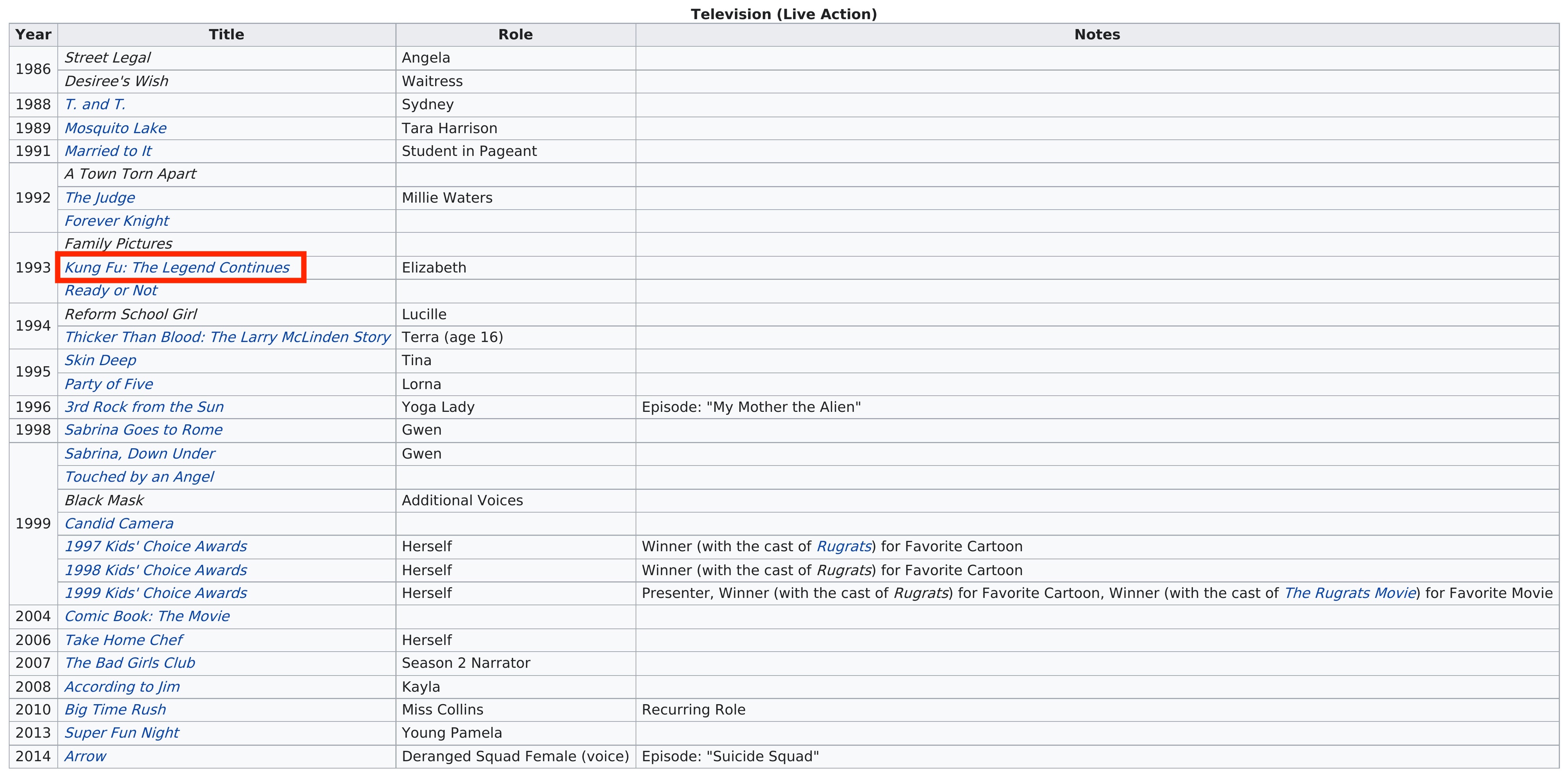}}
	\caption{\textbf{\papertitle predictions on WikiTableQuestions}: For the question \"which movie came out in 1993 and has the role of elizabeth?\", \papertitle correctly predicts \"Kung Fu: The Legend Continues\". We hypothesize that our Token-to-Line pre-training helps in such scenarios as our approach has efficiently learned to make sense of local information (here information in a row). Image file in test: csv\_201\-csv\_23\_0.jpeg}
	\label{fig:wtq_supplemental_viz4}
\end{figure*}

\subsection{WikiTableQuestions Vizualizations}
\label{sec_supple:wikivqa}
We present \papertitle prediction visualizations in this section. See Fig. \ref{fig:wtq_supplemental_viz1}, \ref{fig:wtq_supplemental_viz2}, \ref{fig:wtq_supplemental_viz3}, \ref{fig:wtq_supplemental_viz4}.

\subsection{FUNSD Vizualizations}
\papertitle achieves state-of-the-art performance of 88.89\% F1-score (see Section 4.3 in the main file) on FUNSD \cite{jaume2019} dataset amongst other multi-modal models with same settings. In this sub-section we look at more visualizations by \papertitle on the test-set. One important aspect of this VDU we would like to mention is the OCR is not in human reading-order.

\begin{figure*}
    {\includegraphics[width=1\linewidth]{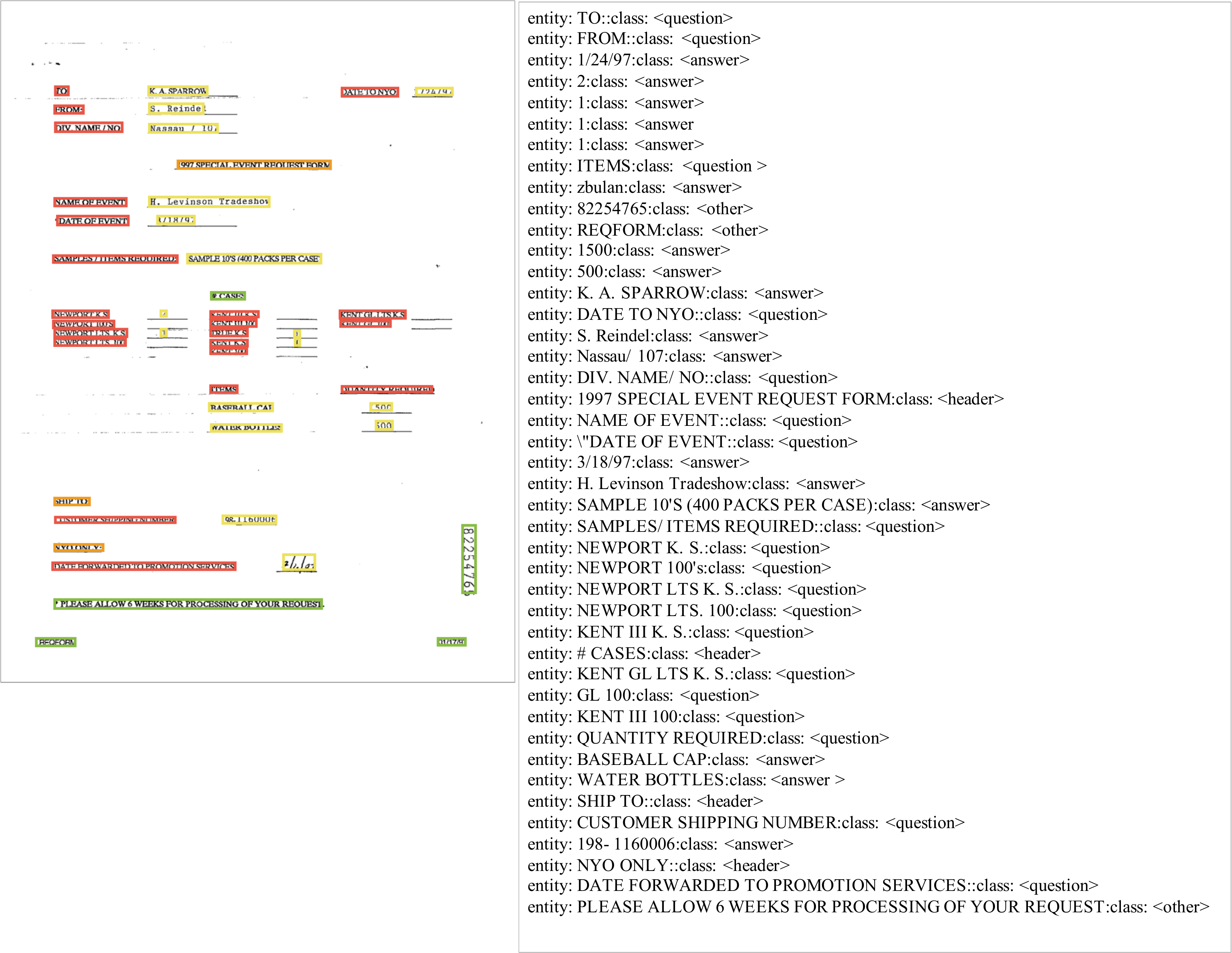}}
	\caption{\textbf{The results of Entity labelling on the sample 82254765 in FUNSD dataset}: Left image shows GT and the right list is the prediction made by \papertitle which correctly match with GT. Best viewed in color. \textcolor{red}{\textbf{Red}} boxes are question entities, \textcolor{yellow}{\textbf{Yellow}} boxes are answer entities, \textcolor{orange}{\textbf{Orange}} boxes are header entities, and \textcolor{green}{\textbf{Green}} boxes are other entities.} 

	\label{fig:funsd_supplemental_viz_pos1}
\end{figure*}
\begin{figure*}
    {\includegraphics[width=1\linewidth]{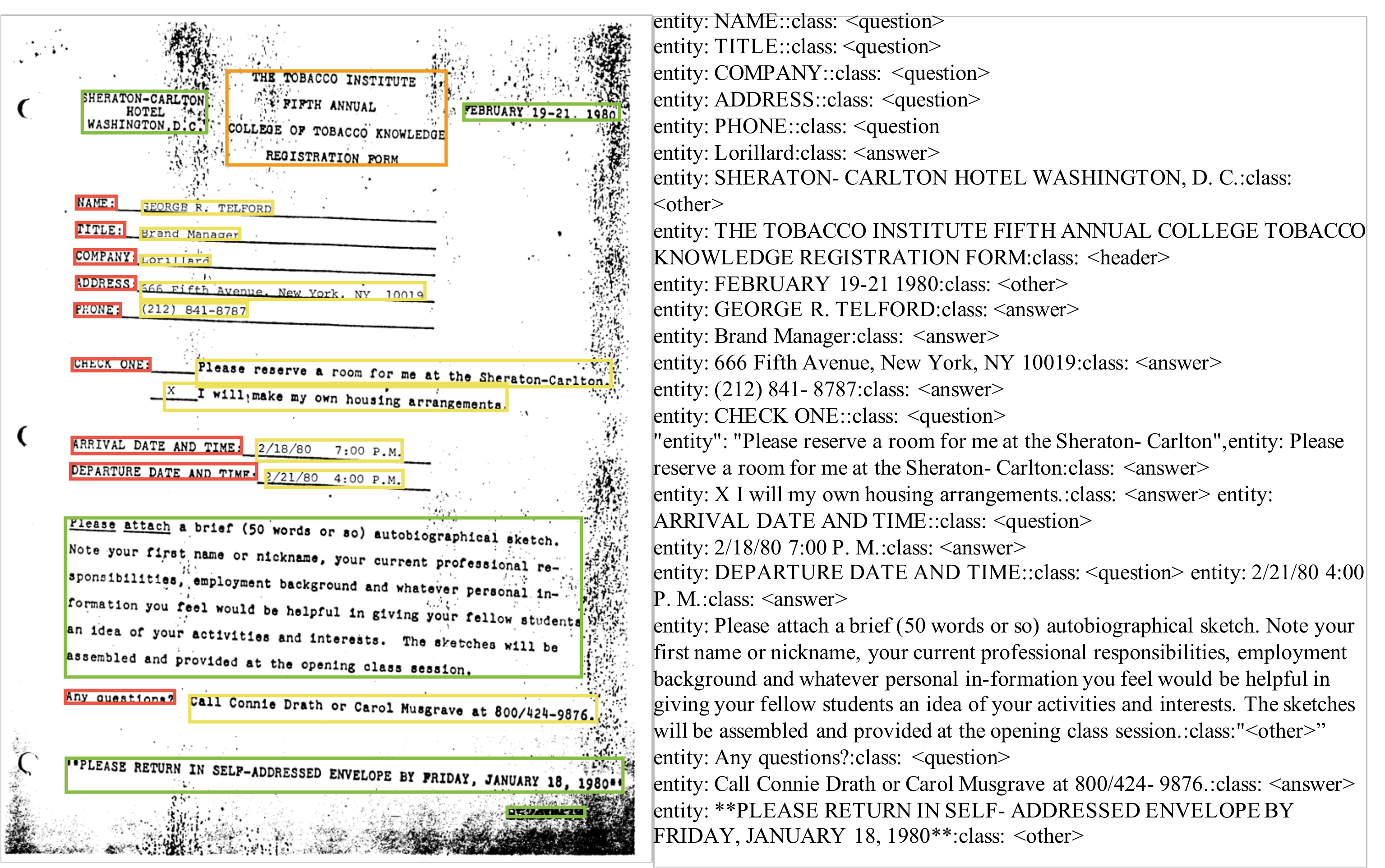}}
	\caption{\textbf{The results of Entity labelling on the sample 85240939 in FUNSD dataset}: Left image shows GT and the right list is the predictions made by \papertitle which perfectly matches with GT. Best viewed in color. \textcolor{red}{\textbf{Red}} boxes are question entities, \textcolor{yellow}{\textbf{Yellow}} boxes are answer entities, \textcolor{orange}{\textbf{Orange}} boxes are header entities, and \textcolor{green}{\textbf{Green}} boxes are other entities.} 

	\label{fig:funsd_supplemental_viz_pos2}
\end{figure*}
\begin{figure*}
	\centering
    {\includegraphics[width=0.9\linewidth]{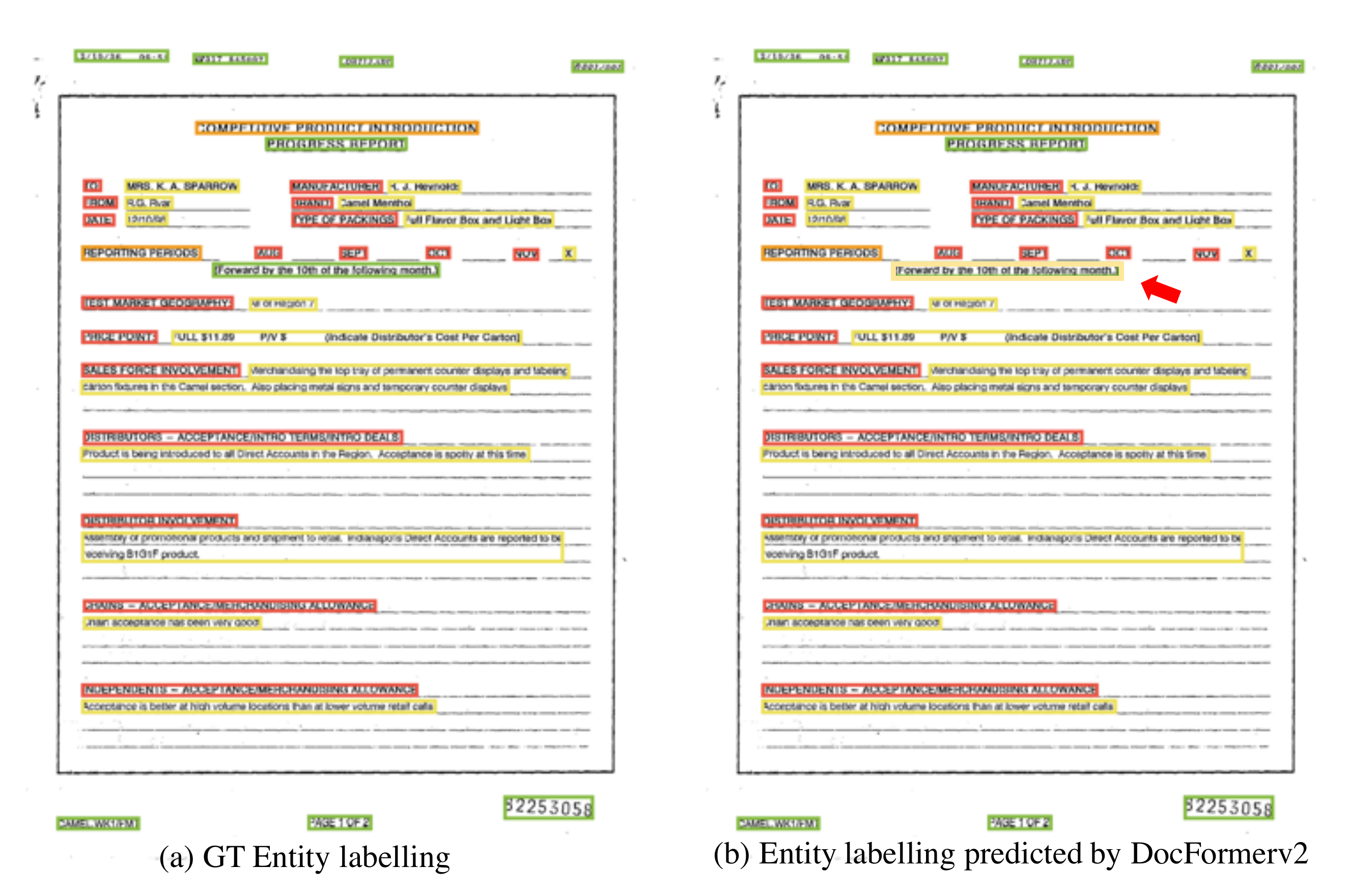}}
	\caption{\small{\textbf{The failure case of entity labelling on FUNSD dataset by \papertitle. }Column(a) shows the ground-truth entity labels, and column (b) is the entities labelling prediction by \papertitle. \textcolor{red}{\textbf{Red}} boxes are question entities, \textcolor{yellow}{\textbf{Yellow}} boxes are answer entities, \textcolor{orange}{\textbf{Orange}} boxes are header entities, and \textcolor{green}{\textbf{Green}} boxes are other entities.} \papertitle wrongly labels ``(Forward by the 10th of the following month.)" as class ``Answer", but its groundtruth class is ``Other".} 

	\label{fig:funsd_supplemental_viz_neg}
\end{figure*}

In Figure \ref{fig:funsd_supplemental_viz_pos1} and \ref{fig:funsd_supplemental_viz_pos2}, we see that \papertitle correctly predicts entity labels on these two samples, and the model can handle the long sequence entity labelling well. We observe some prediction errors on class ``Other' and ``Header', as shown in Figure \ref{fig:funsd_supplemental_viz_neg}. It is mainly due to lack of training samples for these two particular classes.

\subsection{CORD Visualizations}
\begin{figure*}
	\centering
    {\includegraphics[width=0.83\linewidth]{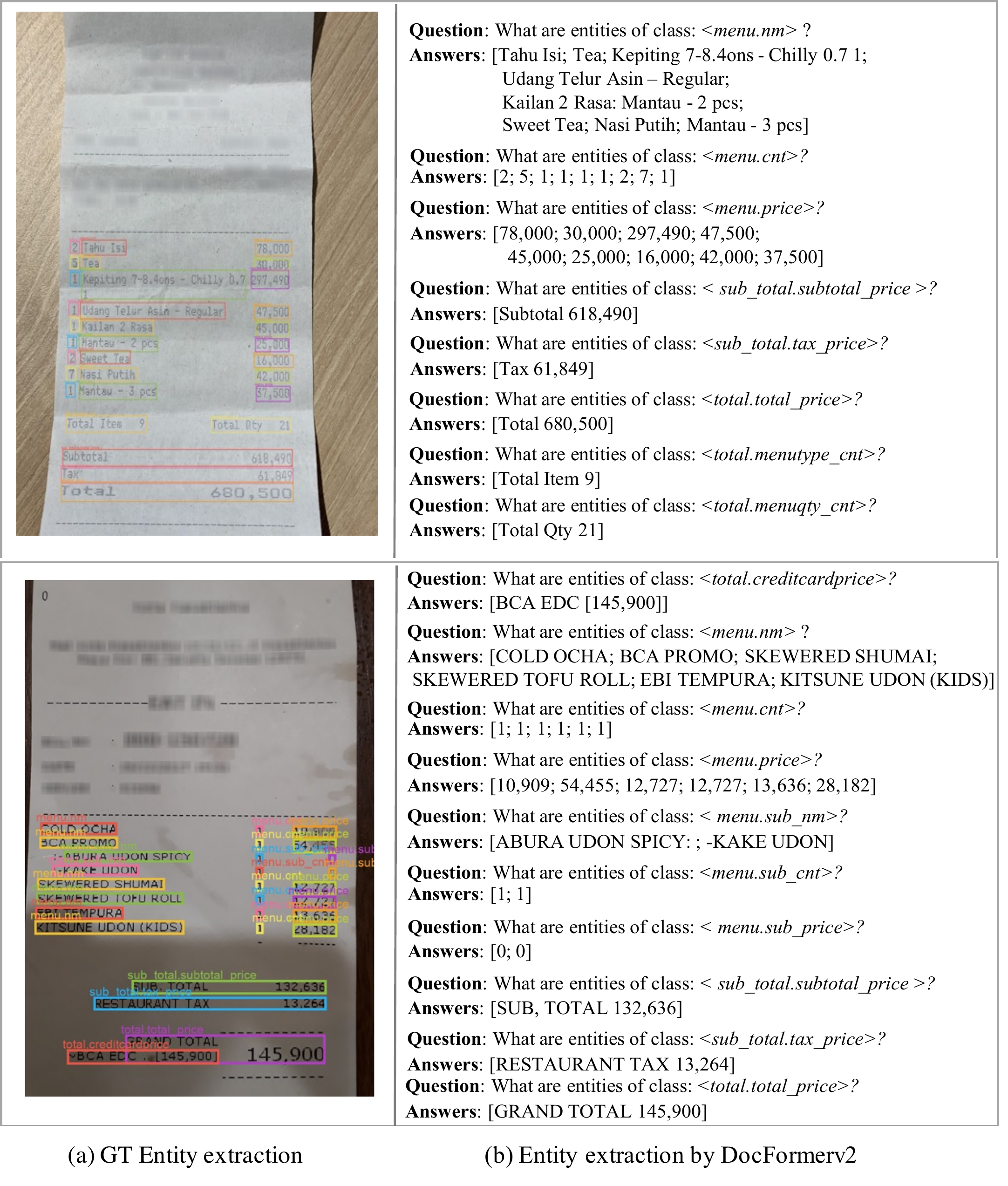}}
	
	\caption{\small{\textbf{Entity extraction comparisons on CORD receipt dataset. }Column(a) shows the Groundtruth entities with the corresponding labels, and column (b) is the entities prediction results by \papertitle. Note that the questions with empty values are not shown. \papertitle predicted correctly all the entity regions in the image. Best if viewed digitally and in color. } } 

	\label{fig:cord_supplemental_viz_pos}
\end{figure*}

\begin{figure*}
	\centering
    {\includegraphics[width=0.83\linewidth]{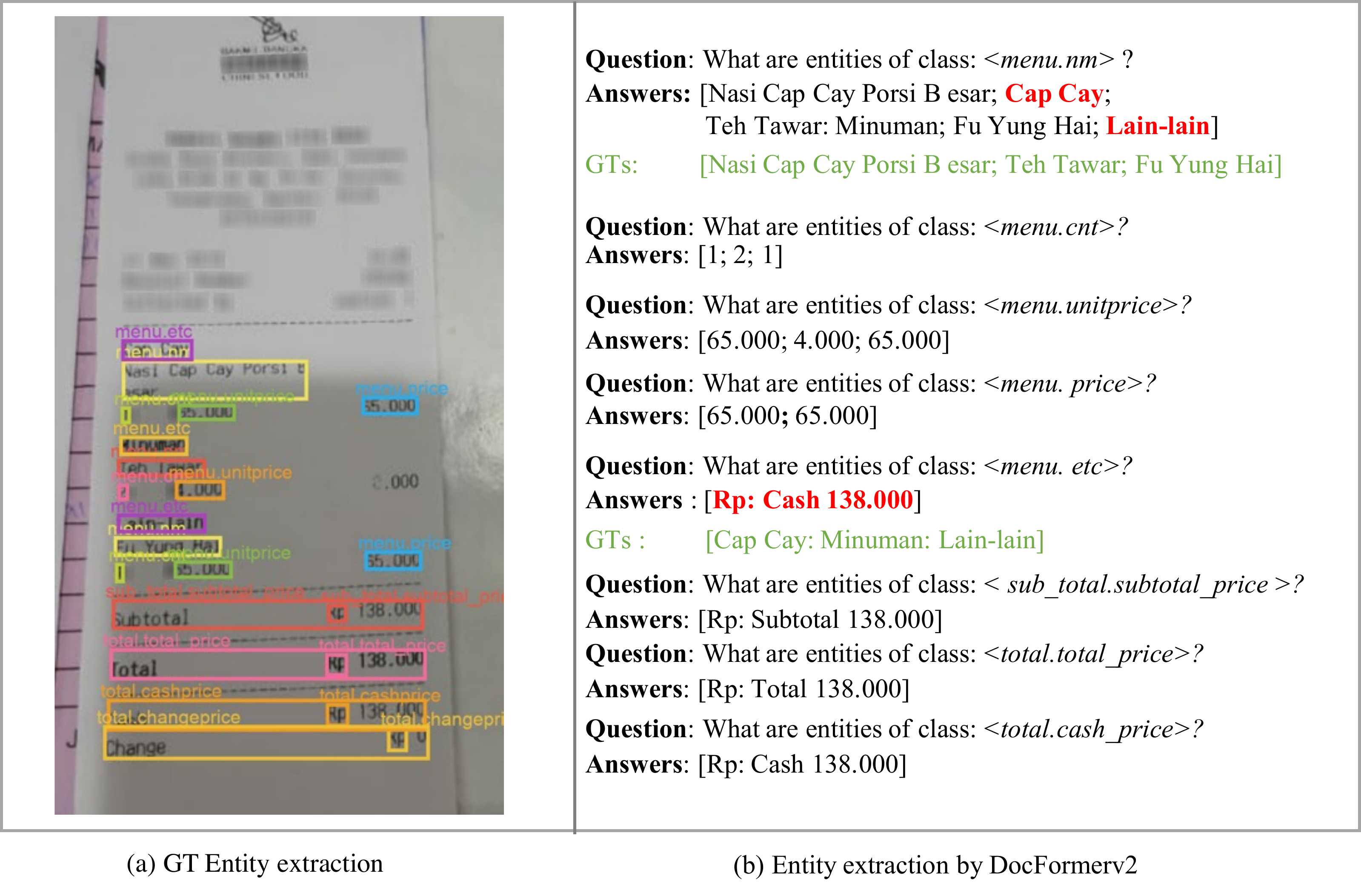}}
	\caption{\small{\textbf{\papertitle Partially correct predictions on CORD dataset. }Column(a) shows the Groundtruth entities with the corresponding labels, and column (b) is the entities prediction results by \papertitle. The entities highlighted in Red are wrongly extracted by \papertitle (The corresponding GT entities are highlighted in green, and blacks entities are correctly extracted). 'Cap Cay' and 'Lain-lain' are wrongly recognized as menu.name. 'Rp: Cash 138,000' is wrongly recognized as menu.etc. It may be because class menu.etc has smaller training samples. Best if viewed digitally and in color. } } 

	\label{fig:cord_supplemental_viz_neg}
\end{figure*}

\papertitle matches the state-of-the-art performance of 97.70\% F1-score on CORD \cite{park2019cord} dataset. Please see Section 4.4 in the main paper. In this sub-section we look at CORD \cite{park2019cord} visualizations by \papertitle. We explicitly show hard-cases where \papertitle does well, see Figures \ref{fig:cord_supplemental_viz_pos}. In the groundtruth visualization, we show the correct entity class on the top-left of the entity boxes. Figure \ref{fig:cord_supplemental_viz_neg} shows some wrong predictions by \papertitle. It is because class
menu.etc has smaller training samples, and the order of the words in this images is different from the reading order.

\subsection{Scene-Text VQA Visualizations}
\begin{figure*}
	\centering
    {\includegraphics[width=0.8\linewidth]{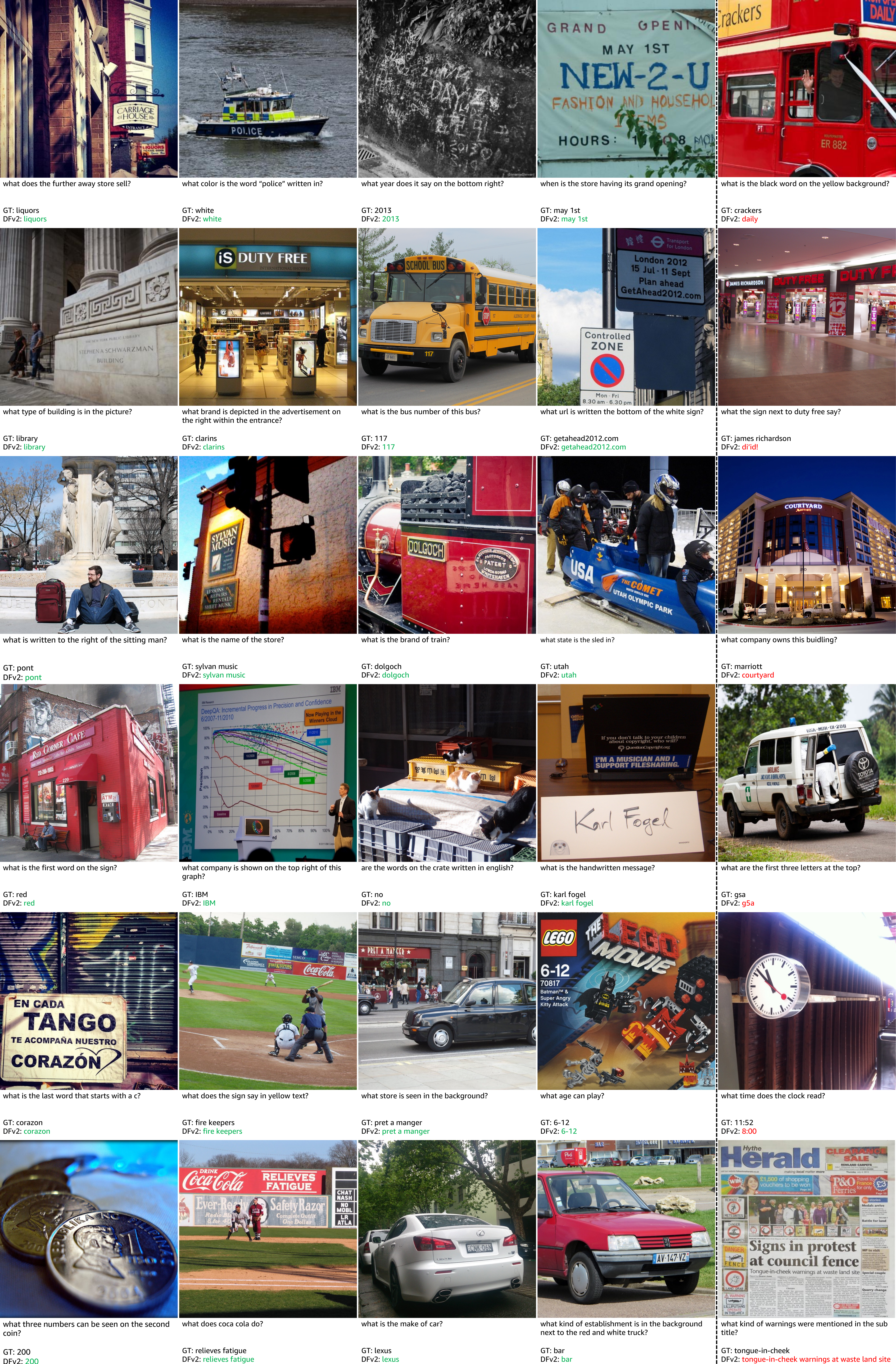}}
	
	\caption{\textbf{\papertitle (DFv2) predictions on Scene-Text VQA}: The first four columns show examples of correct predictions from DocFormerv2. The last column shows examples of failure cases of DocFormerv2.} 

	\label{fig:stvqa_vis}
\end{figure*}

Figure \ref{fig:stvqa_vis} shows examples of correct and wrong predictions from DocFormerv2.
As we can see, \papertitle is able to understand information from different modalities to give correct answers.
For example, for the fourth image in the last row,
the model needs to understand the meanings of ``establishment'' (language), ``next to'' (spatial), ``red'' (visual), \etc to infer the correct answer.
There are multiple reasons for the error cases, including OCR error, hard images/questions, ground truth mismatch, \etc. 
For example, for the last image in the fourth row, \papertitle gives prediction ``g5a'' (\vs ground truth ``gsa''). 
This is because OCR wrongly recognizes the word ``gsa'' as ``g5a''.
For the last image in the fifth row,
it is even difficult for human to figure out the correct answer (we saw big discrepancies in the annotations from different annotators).
For the last image in the last row,
the ground truth is ``tongue-in-cheek warnings'' and \papertitle gives prediction ``tongue-in-cheek warnings at waste land site''.
We can see our prediction is reasonable and just gives more information in the prediction.


\newpage

\FloatBarrier
\clearpage
{\small
\bibliographystyle{ieee_fullname}
\bibliography{egbib}
}